\documentclass[11pt]{article}

\usepackage[final]{acl}

\usepackage{times}
\usepackage{latexsym}

\usepackage[T1]{fontenc}

\usepackage[utf8]{inputenc}

\usepackage{microtype}

\usepackage{inconsolata}

\usepackage{graphicx}

\usepackage{amsmath}
\usepackage{amssymb}
\usepackage{booktabs}
\usepackage{algorithm} 
\usepackage{algorithmic}
\usepackage{multirow}
\usepackage{tabularx}
\usepackage{listings}

\title{VocabTailor: Dynamic Vocabulary Selection for Downstream Tasks in Small Language Models}

\author{
    \textbf{Hanling Zhang\textsuperscript{1,2}}\thanks{Equal contribution.},
    \textbf{Yayu Zhou\textsuperscript{7}}\footnotemark[1],
    \textbf{Tongcheng Fang \textsuperscript{3}},
    \textbf{Zhihang Yuan\textsuperscript{3}}\thanks{Co-corresponding author.}, 
\\
    \textbf{Guohao Dai \textsuperscript{2,5}},
    \textbf{Wanli Ouyang \textsuperscript{1,4,6}},
    \textbf{Yu Wang \textsuperscript{3}}\footnotemark[3]
\\
\\
    \textsuperscript{1}The Chinese University of Hong Kong,
    \textsuperscript{2}Infinigence AI,
    \textsuperscript{3}Tsinghua University, 
\\
    \textsuperscript{4}SLAI,
    \textsuperscript{5}Shanghai Jiao Tong University, 
\\
    \textsuperscript{6}Shanghai Artificial Intelligence Laboratory,
    \textsuperscript{7}Independent Researcher
\\
}

\newcommand{\eg}{e.g.,}
\newcommand{\ie}{i.e.,}

\newtheorem{observation}{Observation}

\newcommand{\set}[1]{\mathcal{#1}} 
\newcommand{\code}[1]{\texttt{#1}} 

\newcommand{\splitlinear}{SplitLinear}
\newcommand{\prealloc}{PreAlloc}
\newcommand{\diskemb}{DiskEmb}

\newif\iffinal
\finaltrue

\iffinal
  \newcommand{\zyyadd}[1]{#1}
  \newcommand{\zyyreplace}[2]{#2}
  \newcommand{\zyydelete}[1]{}

\else
    \usepackage{ulem}
    \newcommand{\stkout}[1]{\ifmmode\text{\sout{\ensuremath{#1}}}\else\sout{#1}\fi} 
    \newcommand{\zyyadd}[1]{{\textcolor{red}{#1}}}
    \newcommand{\zyyreplace}[2]{\textcolor{red}{\stkout{#1}} \textcolor{blue}{#2}}
    \newcommand{\zyydelete}[1]{\textcolor{red}{\stkout{#1}}}

\fi

\begin{document}

\maketitle

\begin{abstract}
Small Language Models (SLMs) provide computational advantages in resource-constrained environments, yet memory limitations remain a critical bottleneck for edge device deployment. A substantial portion of SLMs' memory footprint stems from vocabulary-related components, particularly embeddings and language modeling (LM) heads, due to large vocabulary sizes. Existing static vocabulary pruning, while reducing memory usage, suffers from rigid, one-size-fits-all designs that cause information loss \zyyreplace{from}{during} the prefill stage and \zyydelete{a }lack \zyydelete{of }flexibility. In this work, we identify two key principles underlying the vocabulary reduction challenge: the \emph{lexical locality} principle, the observation that only a small subset of tokens is required during any single inference, and the \emph{asymmetry in computational characteristics} between vocabulary-related components of SLM. Based on these insights, we introduce \textbf{VocabTailor}, a novel decoupled dynamic vocabulary selection framework that addresses memory constraints through offloading embedding and implements a hybrid static-dynamic vocabulary selection strategy for LM Head, enabling on-demand loading of vocabulary components. Comprehensive experiments across diverse downstream tasks demonstrate that VocabTailor achieves a reduction of up to 99\% in the memory usage of vocabulary-related components with minimal or no degradation in task performance, substantially outperforming existing static vocabulary pruning. Our code is available at \url{https://github.com/AwakenedInsects/VocabTailor}.
\end{abstract}

\section{Introduction}\label{sec:intro}
\begin{figure}[ht]
    \centering
    \includegraphics[width=\linewidth]{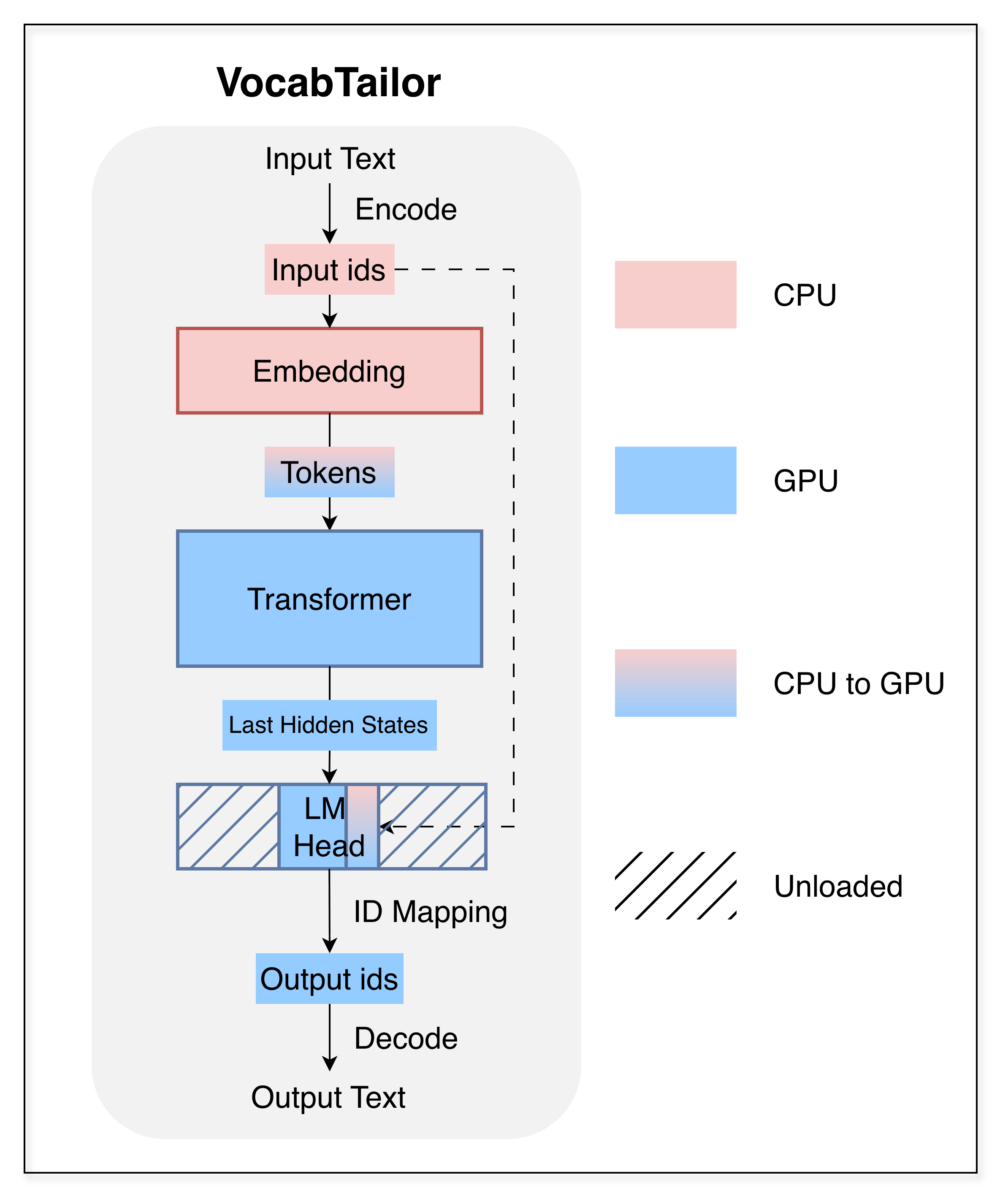}
    \caption{Overview of VocabTailor}
    \label{fig:framework}
\end{figure}

Large Language Models (LLMs) \citep{brown2020language, touvron2023llama, touvron2023llama-1, achiam2023gpt, team2024gemini, anthropic_claude_2024, anthropic_introducing_2025, bai2023qwen, guo2025deepseek} have rapidly become foundational to modern AI applications. Recently, increasing attention has turned towards small language models (SLMs), which are better suited for deployment on edge devices and in resource-constrained environments. Despite their compact size, memory still remains a bottleneck, particularly for edge devices with limited GPU memory. A key driver of this bottleneck is the model's vocabulary size, which directly impacts the memory footprint of both the embedding layer and the language modeling (LM) head. For example, in the Llama 3.2 1B model with a 128K-token vocabulary, the embedding and LM head account for over 20\% of the total memory usage. As SLMs are scaled down and deployed under tighter memory constraints, vocabulary-related memory inefficiencies become increasingly unsustainable, posing a fundamental barrier to efficient SLM deployment.

To address this, prior work has explored static vocabulary pruning strategies \citep{ushio_efficient_2023, yang_textpruner_2022}, which reduce vocabulary size by eliminating rare or irrelevant tokens based on curated corpora. While these approaches are well-motivated, they suffer from key limitations due to their static and coupled design, which assumes a single, globally pruned vocabulary is applied on all vocabulary-related components (\ie\, tokenizer, embedding, and LM head). This design introduces two major issues. First, \textbf{premature information loss} arises because pruning the tokenizer, embedding, and LM head altogether alters the input representation passed to the transformer. These modified inputs may differ significantly from those seen during pretraining, causing distributional shifts and information loss from the prefill stage. Notably, each pruning step in the pipeline introduces information loss that accumulates, leading to cumulative performance degradation during inference. Second, \textbf{lack of flexibility and adaptability} as the static strategy limits the model’s adaptability across diverse tasks. Supporting different task configurations typically requires duplicating multiple copies of vocabulary-related components, resulting in substantial storage overhead and increased complexity in deployment.

Based on empirical observations and theoretical analysis, we derive two key principles for efficient design. \textbf{Lexical locality} captures the empirical observations that, in common downstream tasks, generation relies on a highly localized vocabulary, where each output depends on a small subset of input tokens and a limited set of task-specific tokens. \textbf{Computation asymmetry} reflects the distinct computational characteristics of the embedding and LM head: the embedding layer mainly leverages lookup operations, which are computationally cheap but memory-bandwidth bound, whereas the LM head is compute-intensive with massive matrix multiplication requiring immense floating-point power and thus better suited for GPUs. Existing pruning methods ignore such asymmetry, missing opportunities for system-level optimization.

Guided by these principles, we propose \textbf{VocabTailor}, a flexible and efficient framework for dynamic vocabulary selection. VocabTailor is based on two main pillars. First, we adopt a \textbf{decoupled design} for vocabulary-related components. We retain the full tokenizer and offload the embedding layer to CPU memory. Since the embedding lookup is a memory-intensive operation with $\mathcal{O}(1)$ computational complexity, we can strategically offload it to free up valuable GPU memory, with minimal overhead to the overall system performance. Second, we propose \textbf{hybrid static-dynamic vocabulary selection}: at runtime, we dynamically select and load input-relevant tokens while maintaining a small, static set of task-specific tokens to ensure stable and efficient computation in the LM head.

This design enables substantial memory savings without compromising input fidelity and model generality. Compared to static pruning, which retains the union of all input and output tokens $\left( \bigcup \set{I}_i \right) \bigcup \left( \bigcup \set{O}_i \right)$, VocabTailor only needs $\set{I}_i \bigcup \set{T}$ at inference time, where $\set{I}_i$, $\set{O}_i$ are input and output tokens for example $i$, and $\set{T}$ is a small, fixed task-specific token set. Since $|\bigcup \set{I}_i| \gg |\set{I}_i|$, this leads to substantial memory savings and improved task adaptability.

In summary, our contributions are:
\begin{enumerate}
    \item We present the first systematic analysis of vocabulary management in LLMs through the lens of \emph{lexical locality} and \emph{computation asymmetry}.
    \item We propose \textbf{VocabTailor}, a flexible, memory-efficient, and task-adaptive framework that supports a hybrid static-dynamic vocabulary selection strategy along with an enhanced profiling strategy.
    \item Across five representative downstream tasks---machine translation, summarization, code completion, information extraction, and math problem solving---VocabTailor reduces the memory usage of vocabulary-related components \zyyadd{(\ie\ the embedding matrix and LM head)} by up to 99\%, with minimal or no performance degradation.
\end{enumerate}

\section{Related Work}\label{sec:related_work}
\subsection{Small Language Model}
Small language models (SLMs) are compact alternatives to large language models (LLMs), which are designed for efficiency, lower computational costs, and deployment on resource-constrained devices. While LLMs like GPT-4 \citep{achiam2023gpt}, LLaMA \citep{touvron2023llama,touvron2023llama-1}, Claude \citep{anthropic_claude_2024, anthropic_introducing_2025}, Gemini \citep{team2024gemini}, Qwen \citep{bai2023qwen}, and DeepSeek \citep{guo2025deepseek} have achieved widespread success across real-life applications, SLMs are gaining attention due to their suitability in GPU memory-constrained environments, personal devices, and task-specific scenarios. They offer a scalable, efficient, and sustainable solution tailored for real-time and on-device applications \citep{lamaakal2025tiny}. With careful selection, small open models can rival and even outperform LLMs while offering improved speed and memory efficiency \citep{sinha2024small}.

Despite advances in architectures, training techniques, and model compression techniques, SLMs still face challenges, including trade-offs between model size and accuracy, generalization limitations, and concerns over bias and privacy \citep{van2024survey, lamaakal2025tiny}. A common approach to building SLMs is distilling them from LLMs while retaining the same tokenizer and vocabulary. This typically causes vocabulary-related components (\ie\ embedding and LM Head) to account for a large proportion of the model's total parameters. This makes vocabulary pruning an effective optimization strategy for SLMs.

\subsection{Tokenization}
Tokenization is a fundamental preprocessing step in Natural Language Processing (NLP) that splits text into smaller units called tokens (\eg\ words or characters), which form the input to downstream tasks. Over the years, various tokenization techniques have been developed \citep{sennrich_neural_2016, kudo2018sentencepiece, devlin_bert_2019}. Among these, Byte-Pair Encoding (BPE) \citep{sennrich_neural_2016} has become one of the most widely used. Originally developed for data compression, BPE was adapted to tokenize text by iteratively merging the most frequent adjacent symbol pairs starting from individual characters until a target vocabulary size is reached. This approach enables efficient representation of frequent words with fewer tokens while breaking rare or unseen words into informative subword units. BPE's widespread adoption across transformer architectures has established it as a core component of LLM infrastructure.

State-of-the-art LLMs such as GPT-4 \citep{achiam2023gpt}, LLaMA \citep{touvron2023llama,touvron2023llama-1}, Gemini \citep{team2024gemini}, Claude \citep{anthropic_claude_2024, anthropic_introducing_2025}, and DeepSeek \citep{guo2025deepseek} rely on BPE-based tokenizers to balance vocabulary efficiency and expressiveness. However, the resulting vocabulary sizes are often large, leading to large embedding matrices and LM heads, which increase computational and memory overhead during inference. This scalability bottleneck has motivated research into vocabulary reduction and adaptive tokenization strategies, especially for resource-constrained deployments.

\subsection{Vocabulary Pruning}
Vocabulary pruning has emerged as a key area of research in NLP, particularly for scaling and deploying efficient language models. During the BERT era, interest in pruning surged as researchers explored ways to streamline BERT and other transformer-based models. More recently, with the rise of small language models (SLMs), efficient vocabulary selection has once again become a pressing concern. 

Vocabulary-trimming \citep{ushio_efficient_2023} and TextPruner \citep{yang_textpruner_2022} were proposed to improve efficiency and reduce model size by removing tokens irrelevant to the target language or rarely seen in downstream tasks. Both methods follow a static pruning strategy: they identify language-specific or task-relevant tokens from a curated corpus and prune the vocabulary accordingly. While this process simultaneously reduces the size of the tokenizer, embedding, and LM head, it introduces premature information loss, leading to cumulative performance degradation, and limited flexibility and adaptability, resulting in substantial memory overhead and increased deployment complexity. These two major limitations underscore the need for more flexible and dynamic vocabulary pruning strategies that are task-specific, adaptable, and generalizable across deployment settings. In response, we propose our framework in Section~\ref{sec:method}.

\section{Method}\label{sec:method}
\subsection{VocabTailor Framework} \label{sec:framework_overview}

VocabTailor (shown in Figure~\ref{fig:framework}) decouples vocabulary management across vocabulary-related components. Unlike static vocabulary pruning that uniformly reduces the tokenizer, embedding layer, and LM head, VocabTailor treats each component based on its unique computational and storage properties.

For the \textbf{tokenizer}, we retain the full vocabulary to preserve input fidelity and prevent information loss during the prefill stage. This avoids the cumulative performance degradation caused by pruned tokenizers, which often fail to fully capture the input expressiveness.

For the embedding layer and LM head, despite their similar roles (tokenize and detokenize), a fundamental computation asymmetry exists between them. The \textbf{embedding layer}, which relies on simple lookup operations, is naturally CPU-friendly and incurs negligible runtime overhead when offloaded to CPU memory from high-cost accelerator memory \cite{yu_scaling_2025}. Thus, we retain it fully and offload it to CPU. In resource-constrained deployment settings such as edge devices or platforms with unified CPU–GPU memory, even CPU-resident embeddings may impose non-trivial memory overhead. To address this, VocabTailor supports disk-backed embedding offloading using the Lightning Memory-Mapped Database (LMDB). Embedding vectors are stored as key–value entries and retrieved on demand during inference, preserving full-vocabulary access while further reducing memory footprint. Implementation details and analysis are in Appendix~\ref{appx:lmdb}.

The \textbf{LM head}, which performs compute-intensive matrix multiplications, must remain on GPU for efficient inference. To optimize its memory footprint, we introduce a hybrid static-dynamic vocabulary selection strategy. This approach leverages the lexical locality and significantly reduces memory usage while preserving downstream task performance and enabling flexible, task-specific adaptation without model duplication.

This decoupled architecture addresses the key limitations of existing approaches while providing a theoretically grounded and practically efficient solution for large-scale vocabulary management. The detailed design and implementation of vocabulary selection strategy are presented in Sections~\ref{sec:analysis}--\ref{sec:vocab_selection_static}.

\subsection{Motivation: Analysis of Lexical Locality} \label{sec:analysis}
To investigate the essence of efficient vocabulary selection, we analyze input-output pairs across diverse datasets aligned with downstream tasks. Our empirical analysis reveals two fundamental properties of lexical locality that motivate VocabTailor's dynamic vocabulary selection strategy.

\begin{observation}
    Input-Driven Locality: Common downstream tasks exhibit strong input-output lexical overlap---each output contains a small subset of input tokens. 
\end{observation}

\begin{figure*}[t]
    \centering
    \includegraphics[width=.95\linewidth]{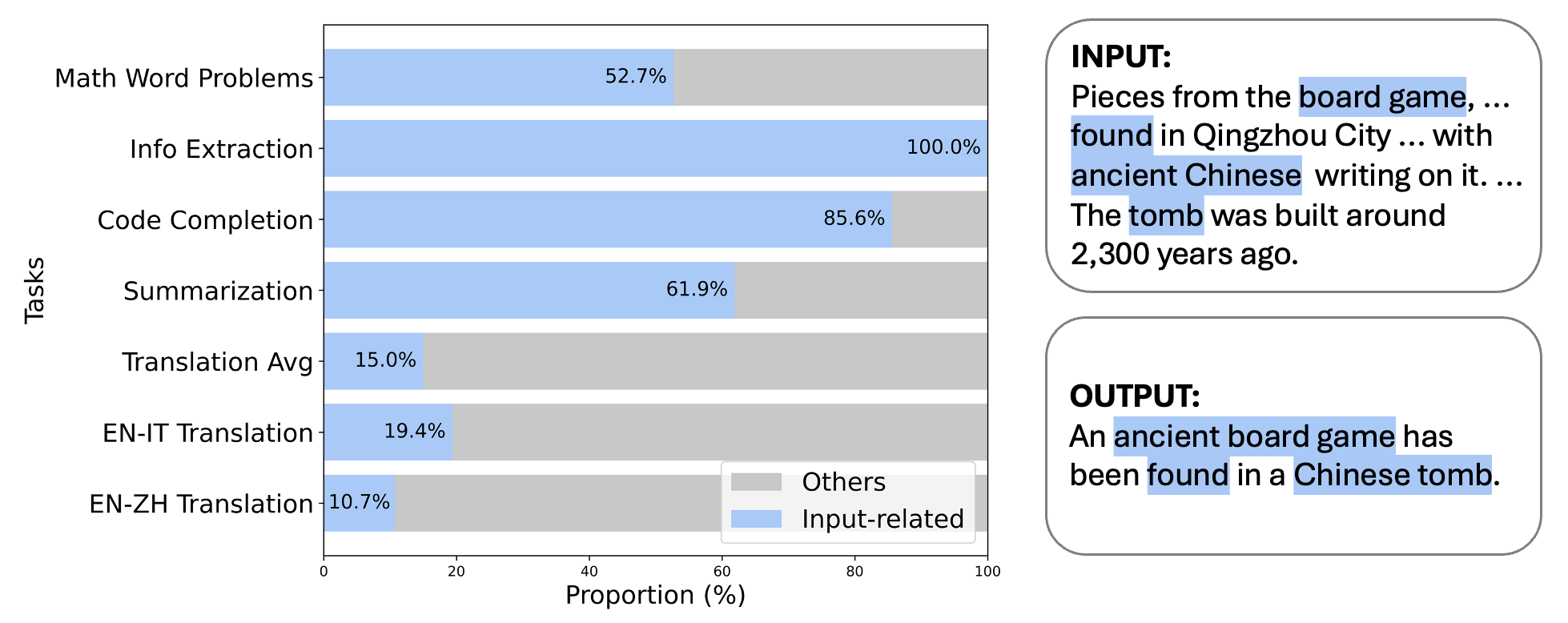}
    \caption{Left: Input-output lexical overlap ratio. Right: Example of lexical overlap in a summarization task.}
    \label{fig:overlap}
\end{figure*}

In various downstream tasks, the model output reuses tokens from the input (shown in Figure~\ref{fig:overlap}). This phenomenon is particularly pronounced in text extraction tasks (e.g., span-based QA or named entity recognition), where output tokens are typically a subset of the input tokens. In code completion, generated code often replicates variable names, function names, and other identifiers from the input context. Similarly, in text summarization, the generated summary contains large spans from the source document. For instance, in summarization, on average 61.9\% of tokens in generated summaries are copied from the input document. From an information-theoretic perspective, the input context dramatically reduces the entropy of the output token distribution, constraining generation to a much smaller effective vocabulary. Thus, preserving input vocabulary is critical for maintaining performance during vocabulary reduction. 

\begin{observation}
    Task-Driven Locality: Vocabulary required for generation is highly localized---each output depends on a limited set of task-specific tokens. 
\end{observation}
Due to input diversity across datasets, the union of all input tokens is significantly larger than the tokens required for any single generation instance. Each input introduces unique tokens, making the aggregate input vocabulary much larger than individual requirements. The remaining output tokens---those not found in the corresponding input---form a relatively small, task-specific set $\mathcal{T}$ that captures the essential generation patterns for the task.

Formally, let $\set{I}$ denote the set of all input tokens, $\set{I}_i$ the set of input tokens for instance $i$ in a dataset, $\set{O}$ the set of all output tokens, and $\set{T} \subset \set{O}$ the set of all task-specific tokens in outputs. We observe:
\begin{equation*}
    |\set{I}| = \left|\bigcup_i \set{I}_i \right| \gg |\set{I}_i| \;\text{and } |\set{O}| > |\set{T}|
\label{eq:obs}
\end{equation*}

These observations expose the fundamental inefficiency of static vocabulary pruning hat operates on the union $\mathcal{I} \cup \mathcal{O}$ for every example. VocabTailor exploits this lexical locality by dynamically constructing active per-example vocabularies using only per-example input tokens $\mathcal{I}_i$ and a compact task-specific set $\mathcal{T}$:
$$\left| \set{I} \bigcup \set{O} \right| \gg \left|\set{I}_i \bigcup \set{T} \right|$$
This dynamic targeting of the much smaller $\mathcal{I}_i \cup \mathcal{T}$ enables substantial memory savings without compromising generation quality, as it only retains the tokens necessary for each inference instance.

\subsection{The Hybrid Vocabulary Selection Strategy} \label{sec:vocab_selection_dynamic}

VocabTailor utilizes lexical locality through a hybrid architecture that combines dynamic runtime adaptation with static offline optimization, enabling efficient vocabulary management without sacrificing generation flexibility.

\subsubsection{Dynamic Selection (Runtime Behavior)}
At the start of each inference instance, VocabTailor identifies the unique input token set $\mathcal{I}_i$ from the input text and selectively loads the corresponding LM head weight vectors from CPU to GPU. This selective loading exploits the input-driven locality principle, ensuring that only input-relevant vocabulary components are active during generation.

Beyond selecting input-related tokens at runtime, VocabTailor must efficiently extend the LM head on the GPU. A naive implementation that repeatedly concatenates static and dynamic weights and instantiates new linear layers incurs unnecessary CPU–GPU transfers and module creation overhead. To mitigate this, VocabTailor supports \prealloc, a pre-allocation strategy that leverages the structure of input-related tokens to amortize LM head construction across inputs.

To minimize the frequency of linear module creation, we investigate the number of unique input-related tokens across different tasks, and we observe that it is relatively small and the distribution is concentrated. Based on this insight, we pre-allocate a small memory space for the dynamic LM head weight. As long as the dynamic weights fit within the buffer size, no concatenation and no new linear module creation are required. The static weights remain on the GPU without movement, so the memory movement overhead is minimized. In the infrequent event that the size of dynamic weights exceeds the current buffer size, a new Linear module with a larger buffer size is created in a way similar to the naive implementation, thereby significantly reducing the expected latency overhead. Implementations and performance evaluation are detailed in Appendix~\ref{appx:dynamic_head_construction}.


\subsubsection{Static Selection (Offline Construction)}
The static component maintains a compact, task-specific core vocabulary $\mathcal{T}$ that captures essential output tokens independent of input context. While simple frequency-based filtering proves inadequate due to noise from typos and multilingual interference in broad corpora, we introduce a theoretically grounded filtering pipeline (Algorithm~\ref{alg:profiling_strategy}) that constructs $\mathcal{T}$ through fine-grained analysis.

\subsection{Fine-grained Construction of the Static Task Vocabulary} \label{sec:vocab_selection_static}
Our static vocabulary construction algorithm (Algorithm~\ref{alg:profiling_strategy}) addresses the challenge of isolating task-essential tokens via a three-stage filtering process:

\noindent\textbf{Input-Aware Filtering.} We first exclude all input tokens from the candidate vocabulary, isolating tokens the model must generate without input cues (e.g., function keywords in code generation, discourse markers in summarization). This step directly implements task-driven locality by identifying the irreducible core $\mathcal{T}$ that cannot be derived from the input context.

\noindent\textbf{Language-Specific Filtering.} To suppress noise in mixed-language scenarios (e.g., code datasets with multilingual comments), we apply Unicode block analysis to retain only tokens belonging to the target language family. This heuristic-based approach effectively handles cross-lingual interference while preserving task-relevant vocabulary.

\noindent\textbf{Tolerance Filtering.} When additional vocabulary reduction is required, we formulate vocabulary selection similar to a pruning problem. We define the impact metric $df(v)$ as the document frequency—the fraction of instances where token $v$ appears in the ground truth. Tokens are sorted by ascending $df(v)$ and iteratively removed until the cumulative document frequency of pruned tokens reaches a user-defined tolerance threshold $\tau$. This threshold bounds the worst-case performance drop, where $\tau = 0.01$ ensures that at most 1\% of profiling samples lose a critical token.

VocabTailor's hybrid approach bridges theoretical rigor (exploiting lexical locality principles) with practical efficiency (optimizing GPU-CPU memory hierarchy). The static-dynamic decomposition enables substantial memory savings while preserving the model's generational capabilities, providing a scalable solution for deploying SLMs across diverse applications with controllable efficiency-accuracy trade-offs.

\section{Experiments} \label{sec:experiments}
\subsection{Settings} \label{sec:exp_settings}
\subsubsection{Tasks and datasets} We evaluate VocabTailor on five representative SLM downstream tasks across diverse domains: machine translation, summarization, code completion, information extraction, and math word problem solving. For machine translation, we involve English-to-Italian and English-to-Chinese translation, as Chinese is logographic with minimal morphology and a large character set, while Italian is alphabetic and morphologically rich. We use the Opus-100 corpus \citep{zhang2020improving} as a profiling dataset and WMT24++ \citep{deutsch2025wmt24++} for the evaluation. For summarization, we use the XSum training set \citep{narayan_dont_2018} to profile and evaluate on its test set. For code completion, the CodeContests$^\text{+}$ corpus \citep{wang2025codecontests+} is used for profiling, while evaluation is conducted on SAFIM \citep{gong2024evaluation}. For information extraction, we use the SQuAD \citep{rajpurkar_squad_2016} training set for profiling and its test set for evaluation. In math problem solving, GSM8K \citep{cobbe2021training} served as the profiling corpora, with evaluation performed on MAWPS \citep{koncel2016mawps}. Each dataset is selected for its strong alignment with the target task. 

\subsubsection{Evaluation metrics}
We include sacreBLEU \citep{post_call_2018}, METEOR \citep{banerjee_meteor_2005}, and COMET \citep{rei_comet_2020} for machine translation. For summarization, we use Rouge-1, Rouge-2, and Rouge-L scores \citep{lin_rouge_2004}. Pass@1 \citep{chen2021evaluating} is used for code completion. We use F1 score for information extraction, and accuracy for math problem solving. These metrics are standard in the field and provide robust measures of model performance across the target tasks. For efficiency evaluation, we report Time to First Token (TTFT), Time Per Output Token (TPOT), Token Per Second (TPS), and Peak VRAM usage. 

\subsubsection{Models}
For machine translation, we use Qwen3-1.7B \citep{yang2025qwen3}. We employ Llama 3.2 3B for summarization and Llama 3.2 1B for information extraction \citep{dubey2024llama}. For summarization, the base model is fine-tuned for a better base performance. For code completion, we choose deepseek-coder-1.3b-base \citep{guo2024deepseek}, and for math problem solving, we apply rho-math-1b-interpreter-v0.1 \citep{lin2024rho}. For efficiency evaluation, we use Qwen3-0.6B \citep{yang2025qwen3}.

\subsubsection{Baselines and Other Settings}
We compare our method (VocabTailor) with the original model (Original) and static vocabulary pruning (VP). Static vocabulary pruning follows the common routines of corpus-based filtering. For a fair comparison, VP and VocabTailor use the same profiling corpra. We set the tolerance threshold $\tau = 0.01$ on all tasks for VocabTailor. As a hybrid dynamic-static framework, the vocabulary size of VocabTailor model varies in each single inference. Here we report the average vocabulary size for each downstream task. For all models, we set the temperature to 0 to avoid randomness. For efficiency evaluation, we use the Qwen3-0.6B model along with 100 prompts from the machine translation (English-to-Chinese) task. We report the latency and VRAM usage on multiple devices, including NVIDIA A100 GPU, Apple Silicon M1 Pro, and Jetson Orin Nano Super. 

\subsection{Results} \label{sec:exp_results}

\begin{table*}[t]
\centering
\begin{tabular}{llrrr}
\toprule
\textbf{Task} & \textbf{Model} & \textbf{Vocabulary} & \textbf{(\% Original)} & \textbf{Metric} \\
\midrule
Machine Translation   & Original    & 151,643            &           & 13.48/12.00/81.16 \\
English$\rightarrow$Chinese                                  & VP          & 64,117             & (42.28\%) & 14.84/11.84/81.19 \\
                                                          & VocabTailor (Ours) & 18,874 + [40]      & (\textbf{12.47\%}) & \textbf{15.39}/\textbf{12.69}/\textbf{81.44} \\
\midrule
Machine Translation   & Original    & 151,643            &           & \textbf{24.33}/\textbf{48.95}/73.87 \\
English$\rightarrow$Italian                                           & VP          & 65,518             & (43.21\%) & 22.59/46.74/74.93 \\
                                                          & VocabTailor (Ours) & 24,185 + [14]      & (\textbf{15.96\%}) & 21.13/46.97/\textbf{75.49} \\
\midrule
Summarization                            & Original    & 128,000            &           & 0.36/0.15/0.29    \\
                                                          & VP          & 59,613             & (46.57\%) & 0.36/0.15/0.29    \\
                                                          & VocabTailor (Ours) & 36,332 + [15]      & (\textbf{28.40\%}) & 0.36/0.15/0.29    \\
\midrule
Code Completion                          & Original    & 32,000             &           & \textbf{54.10\%}             \\
                                                          & VP          & 24,888             & (77.78\%) & 8.12\%              \\
                                                          & VocabTailor (Ours) & 3,521 + [58]       & (\textbf{11.18\%}) & 53.87\%             \\
\midrule
Information Extraction                   & Original    & 128,000            &           & 38.40               \\
                                                          & VP          & 49,106             & (38.36\%) & 2.58                \\
                                                          & VocabTailor (Ours) & [105]          & (\textbf{0.08\%})  & \textbf{62.73}               \\
\midrule
Math Problem Solving                     & Original    & 32,000             &           & 88.40                \\
                                                          & VP          & 10,300             & (32.19\%) & 87.80                     \\
                                                          & VocabTailor (Ours) & 5,135 + [14]             & (\textbf{16.09\%}) & \textbf{88.40}                     \\
\bottomrule
\end{tabular}
\caption{Results on machine translation (sacreBLEU/METEOR/COMET), summarization (Rouge-1/Rouge-2/Rouge-L), code completion (Pass@1), information extraction (F-1), and math problem solving (Accuracy), including the vocabulary size and the ratio to the original model (\%). For VocabTailor, vocabulary size consists of task-specific tokens and the average number of dynamic tokens highlighted in brackets. The best results are in bold characters.}
\label{tab:results}
\end{table*}

As shown in Table~\ref{tab:results}, VocabTailor consistently achieves substantial vocabulary reduction while maintaining or even improving task performance compared with the original model, outperforming static pruning in nearly all cases.

\subsubsection{Qualitative Evaluation}

\noindent\textbf{Machine Translation.} In English-to-Chinese translation, VocabTailor achieves the best results (SacreBLEU 15.39, METEOR 12.69, and COMET 81.44), while using only 12\% of the full vocabulary. These improvements are notable given the inherent difficulty of vocabulary pruning in high-character-set languages like Chinese. \zyyreplace{The results suggest that VocabTailor effectively retains the tokens most essential for both surface-level fluency and deeper semantic adequacy.}{The results suggest that VocabTailor’s dynamic vocabulary selection, which retains input-relevant tokens while filtering low-utility output tokens, preserves both surface-level fluency and semantic adequacy. Trimming irrelevant vocabulary reduces noise in the output distribution and improves decoding stability.}

In English-to-Italian translation, VocabTailor attains the highest COMET (75.49), indicating strong preservation of semantic and contextual meaning. \zyyadd{We attribute this to its ability to retain morphology-relevant tokens while removing unrelated vocabulary, which is particularly beneficial for morphologically rich languages such as Italian.} It retains only 16\% of the original vocabulary, yet performs competitively on SacreBLEU (21.13) and METEOR (46.97), with modest drops compared to the unpruned model. In contrast, VP removes less vocabulary (43\%) but still causes moderate degradation across all metrics, suggesting that static pruning may eliminate rare but task-relevant tokens. VocabTailor’s ability to adapt to Italian’s rich inflectional morphology further reinforces the general applicability of the method.

\noindent\textbf{Summarization.} All three models yield identical ROUGE scores, indicating no difference in summary quality. VocabTailor achieves this with just 28\% of the original vocabulary, 18\% fewer tokens than VP. These results demonstrate that aggressive vocabulary reduction is possible without compromising output quality when applying a hybrid dynamic-static vocabulary selection.

\noindent\textbf{Code Completion.} On the SAFIM benchmark with deepseek-coder-1.3b-base, VP causes a dramatic performance drop: Pass@1 falls from 54.10\% to 8.12\%, despite a modest vocabulary reduction to 78\% of the original size. This highlights a key limitation of corpus-based pruning for code: essential elements such as variable names and identifiers may be discarded if they appear infrequently in the profiling corpus. In contrast, VocabTailor retains only 11\% of the original size and achieves a high Pass@1 of 53.87\%. These findings underscore the robustness of our input- and task-specific vocabulary selection strategy for generation-intensive tasks such as code synthesis.

\begin{table*}[ht]
\centering
\begin{tabular}{llrrrr}
\toprule
\textbf{Model} & \textbf{Device}                  & \textbf{Avg. TTFT} & \textbf{Avg. TPOT} & \textbf{TPS}& \textbf{Peak VRAM} \\
 &                   & (ms) & (ms) & (token/s) & (GB) \\

\midrule
Original       & Apple Silicon M1 Pro             & 23.58             & 75.10      &18.82       & 1.11              \\

       & Jetson Orin Nano Super           & 4.40              & 130.80     & 7.68       & 1.17              \\

       & NVIDIA A100                      & 1.74              & 24.21    & 41.86         & 1.17              \\
\midrule
VocabTailor             & Apple Silicon M1 Pro             & 25.73             & 62.03    & 20.27         & 0.85              \\

             & Jetson Orin Nano Super           & 20.62             & 127.22      & 7.93      & 0.91              \\
             & NVIDIA A100                      & 3.71              & 24.30      & 41.65       & 0.91              \\
\bottomrule
\end{tabular}
\caption{Performance comparison between the Original model and VocabTailor across different devices.}
\label{tab:eff_results}
\end{table*}

\noindent\textbf{Information Extraction.} For information extraction tasks, we evaluate on the SQuAD dataset using Llama 3.2 1B. VocabTailor achieves the most striking result: with just 0.08\% of the original vocabulary retained (and no static tokens at all), it attains an F1 score of 62.73, outperforming both the Original (38.40) and VP (2.58) by large margins. This \zyyreplace{result stems from the nature of}{improvement is expected for} extractive tasks, where the output vocabulary is typically a subset of the input. \zyyreplace{Because VocabTailor preserves input tokens dynamically, it retains all the necessary vocabulary for accurate extraction. VP's poor performance suggests that static, corpus-based pruning is poorly suited for tasks where input-output overlap is high, whereas VocabTailor is especially effective.}{By dynamically preserving all input tokens, VocabTailor guarantees coverage of the necessary output vocabulary, whereas VP may prune such tokens during static filtering. This highlights the limitation of static, corpus-based pruning under high input–output overlap and the advantage of VocabTailor in this setting.} 

\noindent\textbf{Math Word Problem Solving.} The unpruned model achieves a score of 88.40, which is fully preserved under VocabTailor, even though it reduces the vocabulary to just 16\% of the Original. VP, while also reducing vocabulary size to 32\%, causes a slight performance drop to 87.80. This shows that even in tasks requiring high-precision symbolic handling, our approach remains effective.

Across five tasks, VocabTailor consistently demonstrates strong performance while substantially reducing vocabulary size. In many cases, it matches or even exceeds the performance of the unpruned model. Compared to static pruning, which often compromises accuracy, VocabTailor reduces vocabulary more aggressively (up to 99\%\zyyadd{, \eg\ to as low as 0.08\% of the Original in information extraction}) without sacrificing model quality. These findings support that dynamic vocabulary selection is a practical and efficient approach to deploying SLMs in resource-constrained environments.

\subsubsection{Efficiency Evaluation} \label{sec:eff_evaluation}

We evaluate the inference efficiency of VocabTailor compared to the original small language model across diverse hardware platforms, including edge devices (Apple Silicon M1 Pro and Jetson Orin Nano Super) and a high-end GPU (NVIDIA A100). Table~\ref{tab:eff_results} shows that VocabTailor consistently reduces peak VRAM usage by 22-23\% (from 1.11 GB to 0.85 GB \zyyreplace{for}{on} M1 Pro chip and 1.17 GB to 0.91 GB \zyyreplace{for}{on} NVIDIA GPUs), highlighting its effectiveness in alleviating memory bottlenecks \zyydelete{for vocabulary-related components }on resource-constrained devices. In the decoding phase, VocabTailor achieves modest improvements in Time Per Output Token (TPOT) on lower-power hardware (up to 17\% faster on M1 Pro) with negligible impact on the A100, owing to reduced computation in the dynamically trimmed LM head. Time to First Token (TTFT) increases on all devices, due to overhead from on-demand vocabulary loading during the prefill phase. However, as the runtime is dominated by decoding, VocabTailor has comparable TPS to the original model. 

\subsection{Ablation Study} \label{sec:ablation}



To understand the contributions of each component in our framework, we conduct a series of ablation experiments on the SAFIM dataset using the deepseek-coder-1.3b-base model for the code completion task. We primarily focus on evaluating the impact of different vocabulary configurations, including the static and dynamic components, as well as our proposed three-stage filtering process.

\noindent\textbf{Impact of Dynamic vs. Static Vocabulary Components.} We first evaluate the impact of the dynamic and static vocabulary components, both individually and in combination. As shown in Table~\ref{tab:ablation_dynamic_static_parts}, using only the dynamic part that contains tokens profiled from the specific input examples results in a significant performance drop (Pass@1 of 36.06\%). This demonstrates that input tokens alone lack the broader coverage needed for robust code completion. Using only the static part (task-specific tokens) achieves a Pass@1 of 52.30\%. However, this still underperforms the full static-dynamic configuration, which achieves the best result at 53.87\% with nearly the same vocabulary size. This indicates that while the static tokens carry most of the task-relevant capacity, including the dynamic tokens adds critical input-specific nuances, and their combination is essential for optimal performance.

\begin{table}[ht]
\centering
\setlength{\tabcolsep}{1.4pt}
\begin{tabular}{lrr}
\toprule
\textbf{Model} & \textbf{Vocabulary} & \textbf{Pass@1} \\
\midrule
Original               & 100\%      & 54.10\% \\
VP                     & 77.80\%       & 8.12\% \\
Dynamic + Static ($\tau=0.01$) & 11.18\% & 53.87\% \\
Dynamic only           & 0.81\%               & 36.06\% \\
Static only ($\tau=0.01$) & 11.00\%           & 52.30\% \\
\bottomrule
\end{tabular}
\caption{Comparison of the dynamic and static components in VocabTailor.}
\label{tab:ablation_dynamic_static_parts}
\end{table}

We also compare our static-only approach with VP: VP retains 78\% of the original vocabulary---more than our approach---it results in a drastically lower Pass@1 of 8.12\%. This stark contrast underscores a key insight: while both VP and our static-only setup are static, VP's direct modification of the tokenizer and embeddings damages input representations, leading to severe degradation. In contrast, our method retains the full tokenizer and embedding, thereby preserving representational integrity and maintaining high performance even with significantly fewer tokens. 

\noindent\textbf{Impact of Input-aware and Language-specific Filtering.}
As discussed earlier, VocabTailor calibrates the static vocabulary through three-stage filtering. In Table~\ref{tab:ablation_ia_ls}, starting from the unfiltered static vocabulary (78\% of the Original), applying input-aware filtering (IA) reduces the size to 53\% with virtually no performance loss (54.07\% vs. 54.06\%). Adding language-specific filtering (LS) further reduces the vocabulary size to 46\%, while performance slightly improves to 54.09\%. This improvement likely stems from the removal of noisy or irrelevant tokens from multilingual corpora, allowing the model to focus on task-relevant representations. These results demonstrate that IA and LS can significantly compress the vocabulary without degrading accuracy, validating the effectiveness of our static token selection process.

\begin{table}[ht]
\centering
\setlength{\tabcolsep}{1.8pt}
\begin{tabular}{lrr}
\toprule
\textbf{Model} & \textbf{Vocabulary} & \textbf{Pass@1} \\
\midrule
Original                    & 100.00\%    & 54.10\% \\
Dynamic + Unfiltered Static & 77.78\%     & 54.07\% \\
Dynamic + IA                & 52.85\%     & 54.06\% \\
Dynamic + IA + LS           & 45.61\%     & 54.09\% \\
\bottomrule
\end{tabular}
\caption{Comparison of input-aware filtering (IA) and language-specific filtering (LS) in VocabTailor.}
\label{tab:ablation_ia_ls}
\end{table}

\noindent\textbf{Impact of Tolerance Filtering.}
Lastly, we analyze the effect of tolerance filtering, which allows for further reduction of rarely activated tokens based on cumulative document frequency. In Table~\ref{tab:ablation_tolerance}, we vary the tolerance threshold ($\tau$) to observe the trade-off between vocabulary size and accuracy. At $\tau = 0$, we preserve all profiled tokens, achieving a Pass@1 of 54.09\%. As tolerance increases, more tokens are filtered out, reducing the vocabulary to as low as 2.5\% of the original size ($\tau = 0.10$), with a gradual decline in performance. Importantly, even with $\tau = 0.01$, the vocabulary shrinks to 11\% with only 1.8\% drop in Pass@1, suggesting that our method is robust to aggressive reduction. This highlights the flexibility of tolerance as a tuning knob to balance compression vs. accuracy.

\begin{table}[ht]
\centering
\setlength{\tabcolsep}{15.5pt}
\begin{tabular}{lrr}
\toprule
\textbf{Model} & \textbf{Vocabulary} & \textbf{Pass@1} \\
\midrule
$\tau=0$    & 45.61\%     & 54.09\% \\
$\tau=0.01$ & 11.18\%     & 53.87\% \\
$\tau=0.02$ & 7.28\%      & 53.71\% \\
$\tau=0.10$ & 2.50\%      & 52.28\% \\
\bottomrule
\end{tabular}
\caption{Comparison of different tolerance thresholds in VocabTailor.}
\label{tab:ablation_tolerance}
\end{table}

\section{Conclusion} \label{sec:conclusion}
In this paper, we propose a flexible and efficient vocabulary selection framework \zyyreplace{effectively}{to} reduce memory usage during SLM inference. By identifying and leveraging lexical locality \zyyreplace{together with the}{and} computation asymmetry, our method reduce\zyyadd{s} up to 99\% \zyyreplace{in}{of} the memory \zyyreplace{usage of}{associated with} vocabulary-related components \zyyadd{(\ie\ the embedding matrix and LM head) }of SLM while maintain\zyyadd{ing} \zyydelete{the }performance on representative downstream tasks.

\section*{Limitations} \label{sec:limitation}
The proposed framework presented in this paper is only explored on language downstream tasks of SLMs. This method may be extended and applied to VLMs, ALMs, and MLLMs for memory-efficient inference on image/video understanding and audio generation tasks. The method focuses on the memory reduction of SLMs, so (1) although our method mildly increases model decoding speed, inference acceleration is not a goal of this work, and (2) while the method can be applied to LLM, the reduction in memory will be limited.

\section*{Acknowledgments}
This work is supported by the JC STEM Lab of AI for Science and Engineering, funded by The Hong Kong Jockey Club Charities Trust and the Research Grants Council of Hong Kong (Project No. CUHK14213224).

\bibliography{refs}

@techreport{anthropic_claude_2024,
  title = {The {{Claude}} 3 {{Model Family}}: {{Opus}}, {{Sonnet}}, {{Haiku}}},
  shorttitle = {The {{Claude}} 3 {{Model Family}}},
  author = {Anthropic},
  year = {2024},
  month = mar
}

@misc{anthropic_introducing_2025,
  title = {Introducing {{Claude}} 4},
  author = {Anthropic},
  year = {2025},
  month = may,
  howpublished = {https://www.anthropic.com/news/claude-4},
  langid = {english},
  language = {en}
}

@article{bai2023qwen,
  title={Qwen technical report},
  author={Bai, Jinze and Bai, Shuai and Chu, Yunfei and Cui, Zeyu and Dang, Kai and Deng, Xiaodong and Fan, Yang and Ge, Wenbin and Han, Yu and Huang, Fei and others},
  journal={arXiv preprint arXiv:2309.16609},
  year={2023}
}

@article{brown2020language,
  title={Language models are few-shot learners},
  author={Brown, Tom and Mann, Benjamin and Ryder, Nick and Subbiah, Melanie and Kaplan, Jared D and Dhariwal, Prafulla and Neelakantan, Arvind and Shyam, Pranav and Sastry, Girish and Askell, Amanda and others},
  journal={Advances in neural information processing systems},
  volume={33},
  pages={1877--1901},
  year={2020}
}

@article{guo2025deepseek,
  title={Deepseek-r1: Incentivizing reasoning capability in llms via reinforcement learning},
  author={Guo, Daya and Yang, Dejian and Zhang, Haowei and Song, Junxiao and Zhang, Ruoyu and Xu, Runxin and Zhu, Qihao and Ma, Shirong and Wang, Peiyi and Bi, Xiao and others},
  journal={arXiv preprint arXiv:2501.12948},
  year={2025}
}

@article{deutsch2025wmt24++,
  title={Wmt24++: Expanding the language coverage of wmt24 to 55 languages \& dialects},
  author={Deutsch, Daniel and Briakou, Eleftheria and Caswell, Isaac and Finkelstein, Mara and Galor, Rebecca and Juraska, Juraj and Kovacs, Geza and Lui, Alison and Rei, Ricardo and Riesa, Jason and others},
  journal={arXiv preprint arXiv:2502.12404},
  year={2025}
}

@inproceedings{devlin_bert_2019,
  title = {{{BERT}}: {{Pre-training}} of {{Deep Bidirectional Transformers}} for {{Language Understanding}}},
  shorttitle = {{{BERT}}},
  booktitle = {Proceedings of the 2019 {{Conference}} of the {{North American Chapter}} of the {{Association}} for {{Computational Linguistics}}: {{Human Language Technologies}}, {{Volume}} 1 ({{Long}} and {{Short Papers}})},
  author = {Devlin, Jacob and Chang, Ming-Wei and Lee, Kenton and Toutanova, Kristina},
  editor = {Burstein, Jill and Doran, Christy and Solorio, Thamar},
  year = {2019},
  month = jun,
  pages = {4171--4186},
  publisher = {Association for Computational Linguistics},
  address = {Minneapolis, Minnesota},
  doi = {10.18653/v1/N19-1423}
}

@inproceedings{gong2024evaluation,
  title={Evaluation of LLMs on Syntax-Aware Code Fill-in-the-Middle Tasks},
  author={Gong, Linyuan and Wang, Sida and Elhoushi, Mostafa and Cheung, Alvin},
  booktitle={International Conference on Machine Learning},
  pages={15907--15928},
  year={2024},
  organization={PMLR}
}

@article{guo2024deepseek,
  title={DeepSeek-Coder: When the Large Language Model Meets Programming--The Rise of Code Intelligence},
  author={Guo, Daya and Zhu, Qihao and Yang, Dejian and Xie, Zhenda and Dong, Kai and Zhang, Wentao and Chen, Guanting and Bi, Xiao and Li, YK and others},
  journal={arXiv preprint arXiv:2401.14196},
  year={2024}
}

@article{kudo2018sentencepiece,
  title={SentencePiece: A simple and language independent subword tokenizer and detokenizer for neural text processing},
  author={Kudo, Taku and Richardson, John},
  journal={arXiv preprint arXiv:1808.06226},
  year={2018}
}

@article{lamaakal2025tiny,
  title={Tiny language models for automation and control: Overview, potential applications, and future research directions},
  author={Lamaakal, Ismail and Maleh, Yassine and El Makkaoui, Khalid and Ouahbi, Ibrahim and P{\l}awiak, Pawe{\l} and Alfarraj, Osama and Almousa, May and Abd El-Latif, Ahmed A},
  journal={Sensors},
  volume={25},
  number={5},
  pages={1318},
  year={2025},
  publisher={MDPI}
}

@misc{narayan_dont_2018,
  title = {Don't {{Give Me}} the {{Details}}, {{Just}} the {{Summary}}! {{Topic-Aware Convolutional Neural Networks}} for {{Extreme Summarization}}},
  author = {Narayan, Shashi and Cohen, Shay B. and Lapata, Mirella},
  year = {2018},
  month = aug,
  number = {arXiv:1808.08745},
  eprint = {1808.08745},
  primaryclass = {cs},
  publisher = {arXiv},
  doi = {10.48550/arXiv.1808.08745},
  archiveprefix = {arXiv}
}

@article{van2024survey,
  title={A survey of small language models},
  author={Van Nguyen, Chien and Shen, Xuan and Aponte, Ryan and Xia, Yu and Basu, Samyadeep and Hu, Zhengmian and Chen, Jian and Parmar, Mihir and Kunapuli, Sasidhar and Barrow, Joe and others},
  journal={arXiv preprint arXiv:2410.20011},
  year={2024}
}

@article{achiam2023gpt,
  title={Gpt-4 technical report},
  author={Achiam, Josh and Adler, Steven and Agarwal, Sandhini and Ahmad, Lama and Akkaya, Ilge and Aleman, Florencia Leoni and Almeida, Diogo and Altenschmidt, Janko and Altman, Sam and Anadkat, Shyamal and others},
  journal={arXiv preprint arXiv:2303.08774},
  year={2023}
}

@misc{rajpurkar_squad_2016,
  title = {{{SQuAD}}: 100,000+ {{Questions}} for {{Machine Comprehension}} of {{Text}}},
  shorttitle = {{{SQuAD}}},
  author = {Rajpurkar, Pranav and Zhang, Jian and Lopyrev, Konstantin and Liang, Percy},
  year = {2016},
  month = oct,
  number = {arXiv:1606.05250},
  eprint = {1606.05250},
  primaryclass = {cs},
  publisher = {arXiv},
  doi = {10.48550/arXiv.1606.05250},
  archiveprefix = {arXiv}
}

@inproceedings{sennrich_neural_2016,
  title = {Neural {{Machine Translation}} of {{Rare Words}} with {{Subword Units}}},
  booktitle = {Proceedings of the 54th {{Annual Meeting}} of the {{Association}} for {{Computational Linguistics}} ({{Volume}} 1: {{Long Papers}})},
  author = {Sennrich, Rico and Haddow, Barry and Birch, Alexandra},
  editor = {Erk, Katrin and Smith, Noah A.},
  year = {2016},
  month = aug,
  pages = {1715--1725},
  publisher = {Association for Computational Linguistics},
  address = {Berlin, Germany},
  doi = {10.18653/v1/P16-1162}
}

@article{sinha2024small,
  title={Are Small Language Models Ready to Compete with Large Language Models for Practical Applications?},
  author={Sinha, Neelabh and Jain, Vinija and Chadha, Aman},
  journal={arXiv preprint arXiv:2406.11402},
  year={2024}
}

@article{team2024gemini,
  title={Gemini 1.5: Unlocking multimodal understanding across millions of tokens of context},
  author={Team, Gemini and Georgiev, Petko and Lei, Ving Ian and Burnell, Ryan and Bai, Libin and Gulati, Anmol and Tanzer, Garrett and Vincent, Damien and Pan, Zhufeng and Wang, Shibo and others},
  journal={arXiv preprint arXiv:2403.05530},
  year={2024}
}

@article{touvron2023llama,
  title={Llama: Open and efficient foundation language models},
  author={Touvron, Hugo and Lavril, Thibaut and Izacard, Gautier and Martinet, Xavier and Lachaux, Marie-Anne and Lacroix, Timoth{\'e}e and Rozi{\`e}re, Baptiste and Goyal, Naman and Hambro, Eric and Azhar, Faisal and others},
  journal={arXiv preprint arXiv:2302.13971},
  year={2023}
}

@article{touvron2023llama-1,
  title={Llama 2: Open foundation and fine-tuned chat models},
  author={Touvron, Hugo and Martin, Louis and Stone, Kevin and Albert, Peter and Almahairi, Amjad and Babaei, Yasmine and Bashlykov, Nikolay and Batra, Soumya and Bhargava, Prajjwal and Bhosale, Shruti and others},
  journal={arXiv preprint arXiv:2307.09288},
  year={2023}
}

@inproceedings{ushio_efficient_2023,
  title = {Efficient {{Multilingual Language Model Compression}} through {{Vocabulary Trimming}}},
  booktitle = {Findings of the {{Association}} for {{Computational Linguistics}}: {{EMNLP}} 2023},
  author = {Ushio, Asahi and Zhou, Yi and {Camacho-Collados}, Jose},
  editor = {Bouamor, Houda and Pino, Juan and Bali, Kalika},
  year = {2023},
  month = dec,
  pages = {14725--14739},
  publisher = {Association for Computational Linguistics},
  address = {Singapore},
  doi = {10.18653/v1/2023.findings-emnlp.981}
}

@inproceedings{yang_textpruner_2022,
  title = {{{TextPruner}}: {{A Model Pruning Toolkit}} for {{Pre-Trained Language Models}}},
  shorttitle = {{{TextPruner}}},
  booktitle = {Proceedings of the 60th {{Annual Meeting}} of the {{Association}} for {{Computational Linguistics}}: {{System Demonstrations}}},
  author = {Yang, Ziqing and Cui, Yiming and Chen, Zhigang},
  editor = {Basile, Valerio and Kozareva, Zornitsa and Stajner, Sanja},
  year = {2022},
  month = may,
  pages = {35--43},
  publisher = {Association for Computational Linguistics},
  address = {Dublin, Ireland},
  doi = {10.18653/v1/2022.acl-demo.4}
}

@misc{yu_scaling_2025,
  title = {Scaling {{Embedding Layers}} in {{Language Models}}},
  author = {Yu, Da and Cohen, Edith and Ghazi, Badih and Huang, Yangsibo and Kamath, Pritish and Kumar, Ravi and Liu, Daogao and Zhang, Chiyuan},
  year = {2025},
  month = may,
  number = {arXiv:2502.01637},
  eprint = {2502.01637},
  primaryclass = {cs},
  publisher = {arXiv},
  doi = {10.48550/arXiv.2502.01637},
  archiveprefix = {arXiv}
}

@misc{zhang_prompting_2023,
  title = {Prompting {{Large Language Model}} for {{Machine Translation}}: {{A Case Study}}},
  shorttitle = {Prompting {{Large Language Model}} for {{Machine Translation}}},
  author = {Zhang, Biao and Haddow, Barry and Birch, Alexandra},
  year = {2023},
  month = jan,
  number = {arXiv:2301.07069},
  eprint = {2301.07069},
  primaryclass = {cs},
  publisher = {arXiv},
  doi = {10.48550/arXiv.2301.07069},
  archiveprefix = {arXiv}
}

@inproceedings{post_call_2018,
    title = "A Call for Clarity in Reporting {BLEU} Scores",
    author = "Post, Matt",
    booktitle = "Proceedings of the Third Conference on Machine Translation: Research Papers",
    month = oct,
    year = "2018",
    address = "Belgium, Brussels",
    publisher = "Association for Computational Linguistics",
    url = "https://www.aclweb.org/anthology/W18-6319",
    pages = "186--191",
}

@inproceedings{banerjee_meteor_2005,
  title     = {{METEOR}: An Automatic Metric for {MT} Evaluation with Improved Correlation with Human Judgments},
  author    = {Banerjee, Satanjeev  and Lavie, Alon},
  booktitle = {Proceedings of the {ACL} Workshop on Intrinsic and Extrinsic Evaluation Measures for Machine Translation and/or Summarization},
  month     = jun,
  year      = {2005},
  address   = {Ann Arbor, Michigan},
  publisher = {Association for Computational Linguistics},
  url       = {https://www.aclweb.org/anthology/W05-0909},
  pages     = {65--72},
}

@inproceedings{rei_comet_2020,
  title={COMET: A Neural Framework for MT Evaluation},
  author={Rei, Ricardo and Stewart, Craig and Farinha, Ana C and Lavie, Alon},
  booktitle={Proceedings of the 2020 Conference on Empirical Methods in Natural Language Processing (EMNLP)},
  pages={2685--2702},
  year={2020}
}

@inproceedings{lin_rouge_2004,
  title={Rouge: A package for automatic evaluation of summaries},
  author={Lin, Chin-Yew},
  booktitle={Text summarization branches out},
  pages={74--81},
  year={2004}
}

@article{yang2025qwen3,
  title={Qwen3 technical report},
  author={Yang, An and Li, Anfeng and Yang, Baosong and Zhang, Beichen and Hui, Binyuan and Zheng, Bo and Yu, Bowen and Gao, Chang and Huang, Chengen and Lv, Chenxu and others},
  journal={arXiv preprint arXiv:2505.09388},
  year={2025}
}

@article{dubey2024llama,
  title={The llama 3 herd of models},
  author={Dubey, Abhimanyu and Jauhri, Abhinav and Pandey, Abhinav and Kadian, Abhishek and Al-Dahle, Ahmad and Letman, Aiesha and Mathur, Akhil and Schelten, Alan and Yang, Amy and Fan, Angela and others},
  journal={arXiv e-prints},
  pages={arXiv--2407},
  year={2024}
}

@article{zhang2020improving,
  title={Improving massively multilingual neural machine translation and zero-shot translation},
  author={Zhang, Biao and Williams, Philip and Titov, Ivan and Sennrich, Rico},
  journal={arXiv preprint arXiv:2004.11867},
  year={2020}
}

@article{wang2025codecontests+,
  title={CodeContests+: High-Quality Test Case Generation for Competitive Programming},
  author={Wang, Zihan and Liu, Siyao and Sun, Yang and Li, Hongyan and Shen, Kai},
  journal={arXiv preprint arXiv:2506.05817},
  year={2025}
}

@inproceedings{koncel2016mawps,
  title={MAWPS: A math word problem repository},
  author={Koncel-Kedziorski, Rik and Roy, Subhro and Amini, Aida and Kushman, Nate and Hajishirzi, Hannaneh},
  booktitle={Proceedings of the 2016 conference of the north american chapter of the association for computational linguistics: human language technologies},
  pages={1152--1157},
  year={2016}
}

@article{cobbe2021training,
  title={Training verifiers to solve math word problems},
  author={Cobbe, Karl and Kosaraju, Vineet and Bavarian, Mohammad and Chen, Mark and Jun, Heewoo and Kaiser, Lukasz and Plappert, Matthias and Tworek, Jerry and Hilton, Jacob and Nakano, Reiichiro and others},
  journal={arXiv preprint arXiv:2110.14168},
  year={2021}
}

@article{lin2024rho,
  title={Rho-1: Not all tokens are what you need},
  author={Lin, Zhenghao and Gou, Zhibin and Gong, Yeyun and Liu, Xiao and Shen, Yelong and Xu, Ruochen and Lin, Chen and Yang, Yujiu and Jiao, Jian and Duan, Nan and others},
  journal={arXiv preprint arXiv:2404.07965},
  year={2024}
}

@article{chen2021evaluating,
  title={Evaluating large language models trained on code},
  author={Chen, Mark and Tworek, Jerry and Jun, Heewoo and Yuan, Qiming and Pinto, Henrique Ponde De Oliveira and Kaplan, Jared and Edwards, Harri and Burda, Yuri and Joseph, Nicholas and Brockman, Greg and others},
  journal={arXiv preprint arXiv:2107.03374},
  year={2021}
}

\clearpage
\appendix

\section{Alternative Designs for Dynamic LM Head Construction} \label{appx:dynamic_head_construction}
\begin{figure*}[t]
    \centering
    \includegraphics[width=\linewidth]{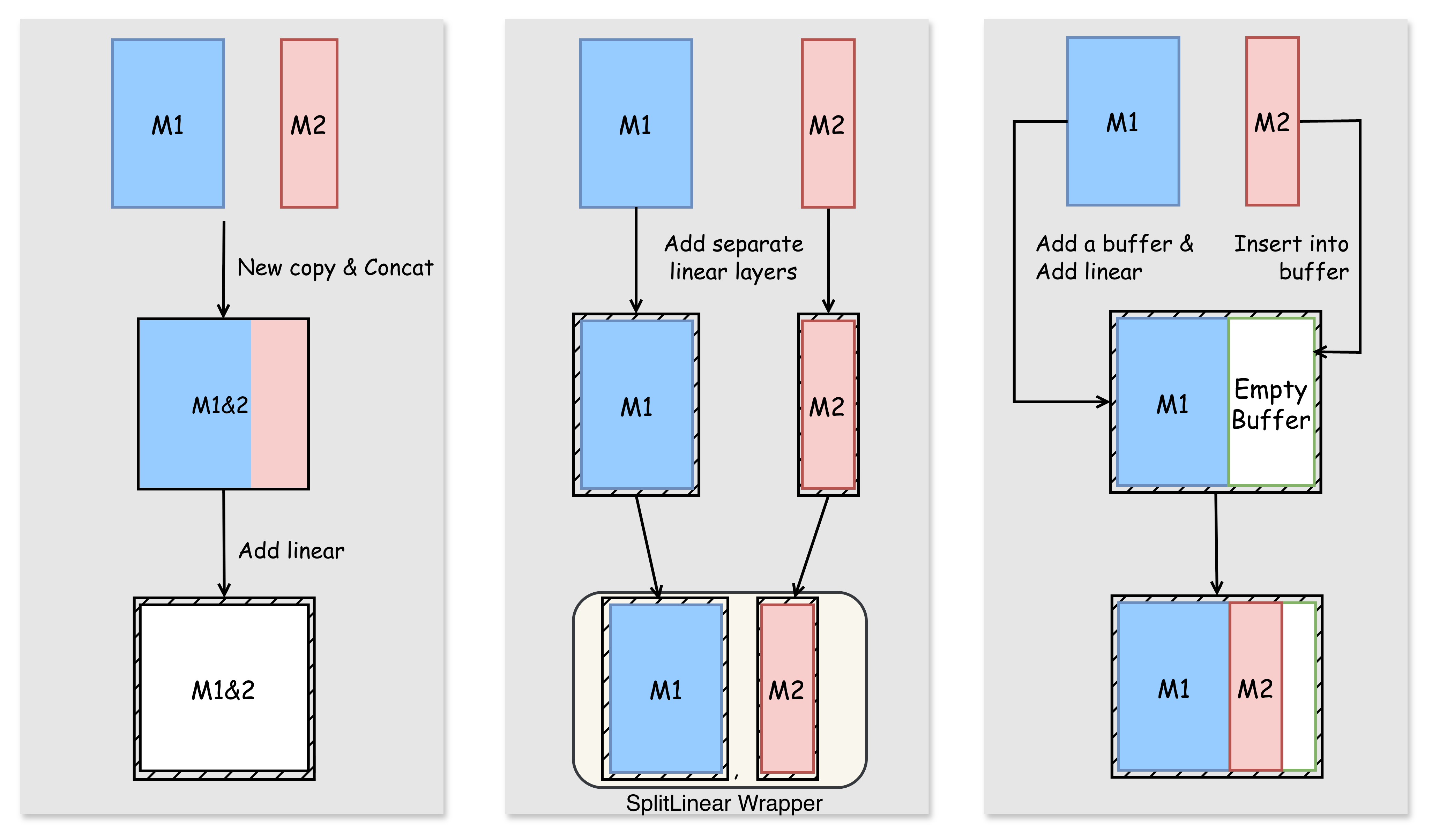}
    \caption{Comparison of dynamic LM head construction workflows. Left panel (Vanilla): The naive re-allocation approach where weights are concatenated and a new Linear module is initialized per request, incurring high movement and initialization costs. Middle panel (\splitlinear): A decoupled architecture where static and dynamic weights form independent Linear modules ($M_1, M_2$), allowing the static part to be pre-initialized on the GPU. Right panel (\prealloc): Our optimized strategy using a pre-warmed GPU buffer. Static weights remain stationary while dynamic weights are copied into a zero-initialized buffer, eliminating module re-initialization and minimizing memory movement overhead.}
    \label{fig:extend_head}
\end{figure*}

The naive (or vanilla) implementation of VocabTailor (Figure~\ref{fig:extend_head}, left panel) dynamically constructs the LM head by selecting the input-related weights, concatenating them with the task-related static weights, onloading the weights from CPU to GPU, and finally creating a new Linear module for the concatenated LM head. This approach has two limitations. First, to avoid the VRAM peak, each time the concatenation needs to be done on the CPU. This results in the frequent CPU-GPU movement of the task-related static part of the LM head weights, which is highly costly for tasks requiring multi-round interactions or multiple calls to SLMs. Second, we need to initialize a new Linear module each time the dynamic LM head is created, and the creation of a linear module introduces extra latency overhead during the prefill stage. To alleviate these latencies and overheads, we propose two alternative approaches: \textbf{\splitlinear} and \textbf{\prealloc}.

\subsection{\splitlinear}
In the \splitlinear\ design (Figure~\ref{fig:extend_head}, middle panel), the static and dynamic parts of the LM head form Linear modules ($M1$ and $M2$) separately. Since the creation of the linear module of the static part can be done in the initialization process, only the creation of the dynamic part needs to be processed during the prefill stage. This design avoids weight concatenation, and the static part of the LM Head can be initialized on the GPU before inference. During the inference process, when the LM head is called, the input is passed through $M1$ and $M2$ independently, and their respective outputs are concatenated to form the final logit. Due to the nature of matrix multiplication, the forward of $M1$ and $M2$ does not interfere with each other, but two GEMMs are required per forward pass.

\subsection{\prealloc}
To minimize the frequency of linear module creation, we investigate the number of unique input-related tokens across different downstream tasks. We observe that the number of unique input-related tokens is relatively small, and the distribution is concentrated. Among all five downstream tasks (machine translation, summarization, code completion, information extraction, and math problem solving), the average number of new unique input-related tokens per request remains below 128. 

Based on this insight, we pre-allocate a small memory space for the dynamic LM head weight. As shown in the right panel of Figure~\ref{fig:extend_head}, we created a Linear module on the GPU in advance, where the weight tensor consists of two parts. The first part is the static weights related to the task, which are loaded before inference. The rest of the weight tensor is a zero-initialized buffer. For each input, the dynamic weights related to the input are copied into the buffer. As long as the dynamic weights fit within the buffer size, no concatenation and no new linear module creation are required. In this process, the static weights remain on the GPU without movement, so the memory movement overhead is minimized. In the infrequent event that the size of dynamic weights exceeds the current buffer size, a new Linear module with a larger buffer size is created in a way similar to VocabTailor's naive implementation, thereby significantly reducing the expected latency overhead.

\subsection{Comparison of Dynamic LM Head Construction Strategies}
Both strategies are included as inference options in VocabTailor because they offer different advantages and disadvantages tailored to various downstream tasks. The \splitlinear\ approach is well-suited for downstream tasks with diverse input-related vocabularies but relatively short outputs, such as the information retrieval task, because it provides instant updates and consistent latency regardless of the number of dynamic input-related tokens. However, multiple GEMMs are performed in the forward pass. Conversely, the \prealloc\ approach is ideal for downstream tasks with relatively small and stable input-related vocabularies, such as code completion or math problem solving, where the output is primarily composed of task-related terms and only a small portion of dynamic input-related vocabularies like variable names. In such scenarios, pre-allocation effectively eliminates both memory movement and module initialization overhead.



\subsubsection{Experimental Setup} 
We test our proposed methods in the machine translation (English-to-Chinese) task using a subset of 100 examples from the WMT24++ dataset \cite{deutsch2025wmt24++}. We use the Qwen3-0.6B \cite{yang2025qwen3} as the base model. All experiments are conducted on a single NVIDIA A100 GPU. We compared four configurations: (1) Original, the original model with standard full-vocabulary; (2) VT (Vanilla), the VocabTailor naive implementation with dynamic CPU-GPU weight concatenation; (3) VT (\splitlinear), which utilizes dual GEMM operations to bypass concatenation; and (4) VT (\prealloc), which employs a pre-warmed GPU buffer of size 128. Performance is measured across 100 prompts to capture statistical distributions of latency metrics including Time to First Token (TTFT), Time Per Output Token (TPOT), and end-to-end (E2E) latency. We also report throughput metrics including Token Per Second (TPS), Request Per Second (RPS), and average output length. Memory usage is monitored via the PyTorch CUDA memory management interface to report peak VRAM consumption.

\subsubsection{Results}
Tables \ref{tab:latency_dist} and \ref{tab:throughput_stats} compare the original model against three VocabTailor (VT) variants, revealing clear trade-offs between initialization overhead, steady-state decoding efficiency, and memory utilization.

\begin{table*}[t]
\centering
\small
\setlength{\tabcolsep}{5.5pt}
\begin{tabular}{lrrrrrrrrrrrr}
\toprule
\multirow{2}{*}{\textbf{Model}}
& \multicolumn{4}{c}{\textbf{TTFT (ms)}} 
& \multicolumn{4}{c}{\textbf{TPOT (ms)}}
& \multicolumn{4}{c}{\textbf{E2E Latency (s)}}\\
\cmidrule(lr){2-5} \cmidrule(lr){6-9} \cmidrule(lr){10-13}
& Mean & P50 & P90 & P99
& Mean & P50 & P90 & P99
& Mean & P50 & P90 & P99 \\
\midrule
Original & 1.74 & 0.79 & 0.90 & 2.07 & 24.21 & 23.64 & 24.09 & 25.17 & 1.57 & 1.56 & 2.44 & 4.12 \\
VT (Vanilla) & 124.29 & 123.22 & 124.63 & 137.43 & 23.99 & 23.47 & 23.60 & 24.74 & 1.66 & 1.66 & 2.47 & 3.91 \\
VT (\splitlinear) & 123.86 & 123.38 & 124.55 & 129.62 & 24.52 & 24.05 & 24.19 & 28.87 & 1.69 & 1.69 & 2.53 & 3.98 \\
VT (\prealloc) & 3.71 & 2.63 & 3.35 & 17.20 & 24.30 & 23.82 & 24.32 & 25.96 & 1.56 & 1.57 & 2.38 & 3.91 \\
\bottomrule
\end{tabular}
\caption{Latency distribution for different dynamic LM head construction approaches on a machine translation (English-to-Chinese) task ($N=100$). We report Time to First Token (TTFT), Time Per Output Token (TPOT), and end-to-end (E2E) latency. Percentiles (Mean, P50, P90, P99) are included to characterize tail latency and model stability under varying prompt complexity.}
\label{tab:latency_dist}
\end{table*}

\begin{table*}[t]
\centering
\begin{tabular}{lrrrr}
\toprule
\multirow{2}{*}{\textbf{Model}}
& \textbf{TPS} & \textbf{RPS} & \textbf{Avg. Output} & \textbf{Peak VRAM} \\
& (token/s) & (req/s) & \textbf{Length} & (GB) \\
\midrule
Original & 41.86 & 0.64 & 65.73 & 1.17 \\
VT (Vanilla) & 42.31 & 0.56 & 64.84 & 0.91 \\
VT (\splitlinear) & 41.34 & 0.55 & 64.84 & 0.91 \\
VT (\prealloc) & 41.65 & 0.55 & 64.84 & 0.91 \\
\bottomrule
\end{tabular}
\caption{Throughput and resource utilization for different dynamic LM head construction approaches on a machine translation (English-to-Chinese) task ($N=100$). Decode throughput is measured in Tokens Per Second (TPS), computed as total generated tokens divided by total decoding time. Request throughput (RPS), average output length, and peak GPU memory consumption are also reported.}
\label{tab:throughput_stats}
\end{table*}

\noindent\textbf{Latency behavior.} The Original achieves the lowest Time to First Token (TTFT), with a mean of 1.74 ms, reflecting its fully static vocabulary and absence of runtime adaptation. In contrast, the Vanilla VT incurs a substantial TTFT increase (mean 124.29 ms), which can be attributed to dynamic vocabulary construction and associated runtime bookkeeping. Notably, this overhead is largely confined to the prefill phase: Time Per Output Token (TPOT) remains comparable across all models ($\approx$24 ms), indicating that VT does not degrade steady-state decoding once generation begins.

Among different dynamic head construction approaches, \prealloc\ is particularly effective at mitigating TTFT overhead, reducing mean TTFT to 3.71 ms, within the same order of magnitude as Original, while preserving the benefits of VT. Tail latency analysis further supports this observation: \prealloc\ substantially improves P99 TTFT (17.20 ms vs. 137.43 ms for Vanilla VT), suggesting improved stability under varying prompt complexities. In contrast, \splitlinear\ does not materially reduce TTFT, as it still requires the dynamic creation of the module $M2$.

\noindent\textbf{End-to-end latency.} Despite large TTFT differences, end-to-end (E2E) latency remains broadly similar across models. All VT variants exhibit mean E2E latency within 1.56--1.69 s, comparable to the Original (1.57 s). This indicates that, for typical MT tasks with non-trivial output lengths, TTFT overhead is amortized over decoding, and overall user-perceived latency is dominated by generation rather than initialization.

\begin{figure*}[ht]
    \centering
    \includegraphics[width=\linewidth]{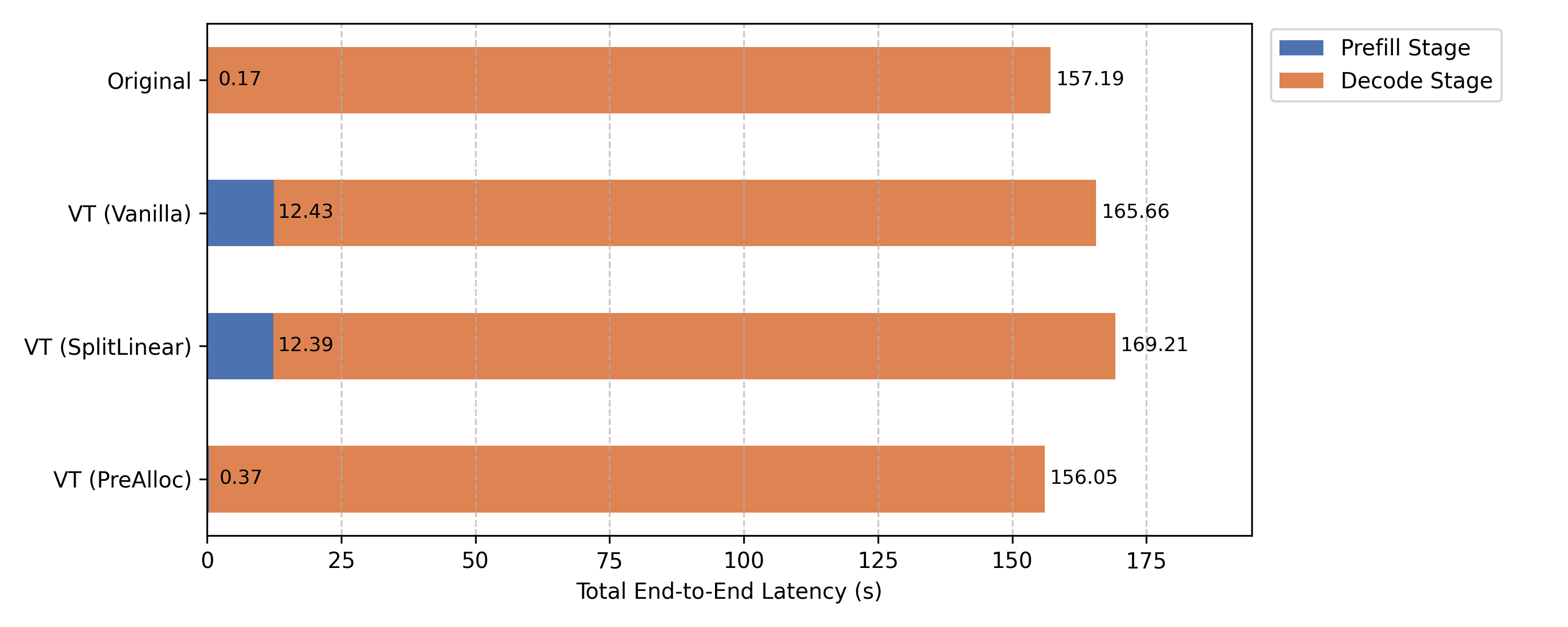}
    \caption{Latency decomposition across different dynamic LM head construction approaches. \prealloc\ (bottom) successfully reduces the total prefill time from 12.43s (Vanilla) back to 0.37s, effectively matching the latency profile of the original model.}
    \label{fig:latency_decomp}
\end{figure*}

\noindent\textbf{Latency decomposition.}
Figure~\ref{fig:latency_decomp} illustrates the breakdown of total E2E latency consumed by the prefill versus decoding stages. In the Vanilla and \splitlinear\ implementations, the prefill stage---which is effectively instantaneous in the baseline ($0.11\%$)---surges to over $7.3\%$ of the total execution time. By utilizing a pre-warmed GPU buffer, the pre-allocation strategy successfully reduce the prefilling time, returning the prefill stage to just $0.24\%$ of total latency.

\noindent\textbf{Throughput and resource efficiency.}
As shown in Table \ref{tab:throughput_stats}, decode throughput (TPS) remains effectively consistent across all models (41--42 tokens/s), confirming that VT and its extensions do not introduce steady-state performance regressions. Request-level throughput, measured by Request Per Second (RPS), is slightly lower for VT-based models, which aligns with their increased per-request initialization cost. In terms of memory footprint, VT variants consistently reduce peak VRAM usage (0.91 GB vs. 1.17 GB for Original), validating the effectiveness of vocabulary decoupling and offloading. 


\noindent\textbf{Overall trade-offs.} Taken together, these results highlight that VocabTailor introduces a clear TTFT–memory trade-off: substantial VRAM savings with minimal impact on decode throughput, at the cost of higher initialization latency. Among the different head construction strategies, \prealloc\ offers the most favorable balance, largely eliminating TTFT penalties while preserving VRAM savings and decode efficiency. \splitlinear\, while maintaining comparable throughput and memory characteristics, provides limited benefit for latency-sensitive scenarios.

\section{Offload Embedding Lookup to LMDB} \label{appx:lmdb}
\begin{figure}[ht]
    \centering
    \includegraphics[width=\columnwidth]{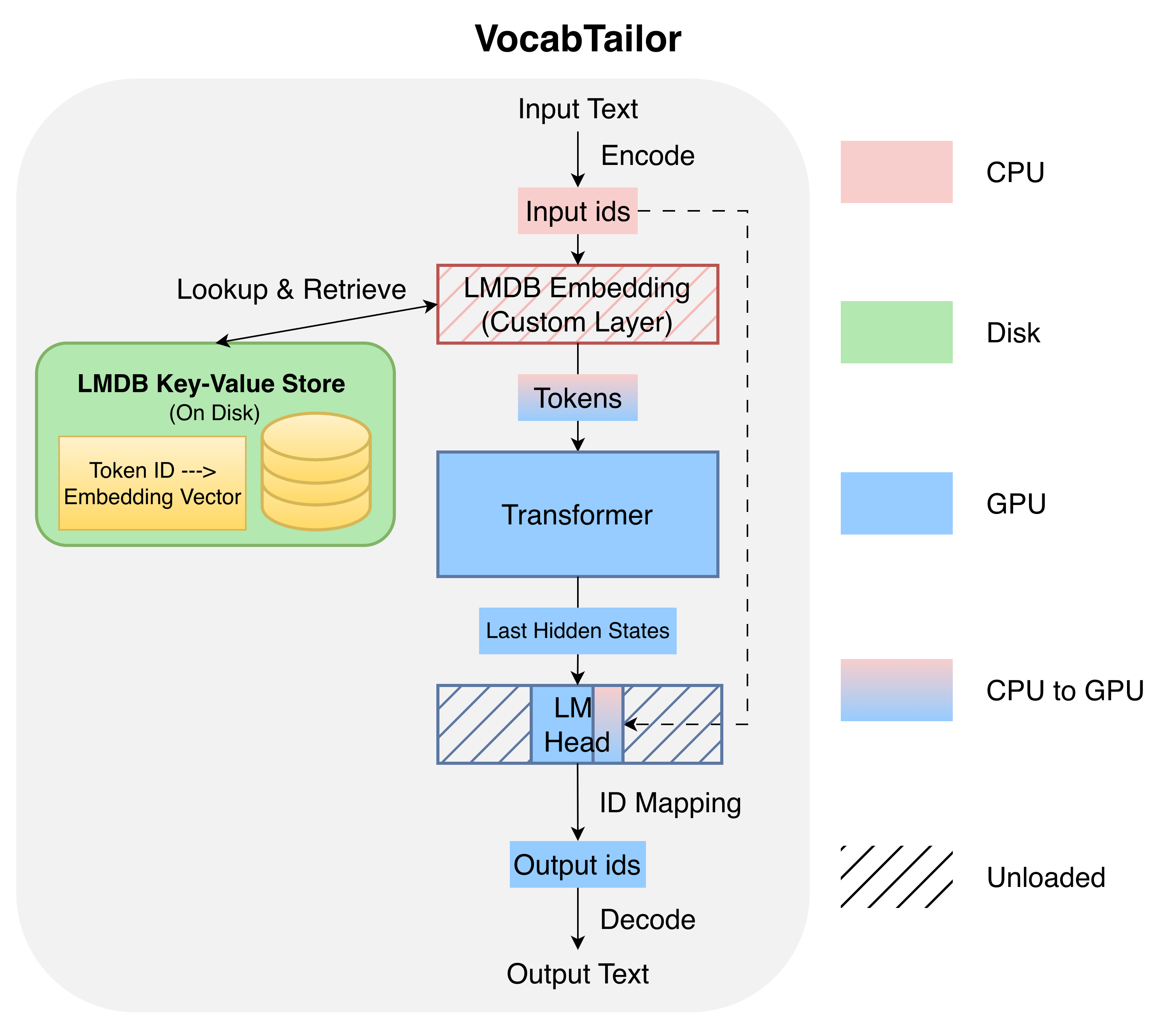}
    \caption{Overview of the VocabTailor framework with disk-backed embedding offloading. The embedding layer is replaced with a custom embedding layer that retrieves the corresponding tokens from the LMDB when the forward is called.}
    \label{fig:framework_lmdb}
\end{figure}

In VocabTailor, model embeddings are offloaded to the CPU to reduce GPU memory usage. However, on devices with limited CPU memory or unified CPU-GPU memory architectures, this design alone does not sufficiently mitigate memory consumption. Such constraints are prevalent in edge devices, where SLMs are frequently deployed, as matrix multiplications can still be executed on the CPU within acceptable latency. In this context, VocabTailor's core principle---decoupling and offloading embedding weights to a lower storage hierarchy---remains effective for two primary reasons: (1) embedding lookup operations are computationally lightweight, and (2) the dominant inference latency arises from the forward passes, particularly tensor multiplications, making data movement a secondary bottleneck. Building on this principle and inspired by prior work on memory-mapped embedding storage~\citep{yu_scaling_2025}, we implement disk-based embedding offloading using the Lightning Memory-Mapped Database (LMDB). The embedding weights are serialized as key-value pairs, with token indices as keys and the corresponding embedding vectors as values. A disk-backed LMDB database is constructed to store these pairs, enabling on-demand retrieval of embeddings during inference. The extended framework is illustrated in Figure~\ref{fig:framework_lmdb}.

\subsection{Experimental Setup} 
We evaluate the effect of embedding offloading for the machine translation (English-to-Chinese) task on a subset of 100 examples from the WMT24++ dataset \cite{deutsch2025wmt24++}. We use the Qwen3-0.6B \cite{yang2025qwen3} as the base model. All experiments are conducted on a single NVIDIA A100 GPU. Building on our previous results identifying \prealloc\ as the most efficient dynamic LM head construction strategy, we compare three configurations that progressively incorporate embedding offloading: (1) Original, the original model with standard full-vocabulary; (2) VT (\prealloc), which offloads embedding weights to CPU and employs a pre-warmed GPU buffer of size 128; and (3) VT (\prealloc) + \diskemb, which further offloads embeddings to disk-backed LMDB storage. Performance is measured across 100 prompts to capture statistical distributions of latency metrics including Time to First Token (TTFT), Time Per Output Token (TPOT), and end-to-end (E2E) latency. We also report throughput metrics including Token Per Second (TPS), Request Per Second (RPS), and average output length. Memory usage is monitored via the PyTorch CUDA memory management interface to report peak VRAM consumption.

\subsection{Results}
Table~\ref{tab:latency_dist_lmdb} presents a comparison between the three configurations. The latter two configurations share the same dynamic LM head construction strategy and differ only in how embedding weights are stored and accessed.

\begin{table*}[t]
\centering
\small
\setlength{\tabcolsep}{5.5pt}
\begin{tabular}{lrrrrrrrrrrrr}
\toprule
\multirow{2}{*}{\textbf{Model}} 
& \multicolumn{4}{c}{\textbf{TTFT (ms)}} 
& \multicolumn{4}{c}{\textbf{TPOT (ms)}}
& \multicolumn{4}{c}{\textbf{E2E Latency (s)}}\\
\cmidrule(lr){2-5} \cmidrule(lr){6-9} \cmidrule(lr){10-13}
& Mean & P50 & P90 & P99
& Mean & P50 & P90 & P99
& Mean & P50 & P90 & P99 \\
\midrule
Original & 1.74 & 0.79 & 0.90 & 2.07 & 24.21 & 23.64 & 24.09 & 25.17 & 1.57 & 1.56 & 2.44 & 4.12 \\
VT (\prealloc) & 3.71 & 2.63 & 3.35 & 17.20 & 24.30 & 23.82 & 24.32 & 25.96 & 1.56 & 1.57 & 2.38 & 3.91 \\
VT (\prealloc) + \diskemb\ & 4.58 & 3.64 & 4.10 & 12.23 & 24.46 & 24.01 & 24.27 & 25.50 & 1.57 & 1.58 & 2.39 & 3.82 \\
\bottomrule
\end{tabular}
\caption{Latency distribution for three inference configurations on a machine translation (English-to-Chinese) task ($N=100$), comparing the effect of embedding offloading. We report Time to First Token (TTFT), Time Per Output Token (TPOT), and end-to-end (E2E) latency. Percentiles (Mean, P50, P90, P99) are included to characterize tail latency and model stability under varying prompt complexity.}
\label{tab:latency_dist_lmdb}
\end{table*}

\begin{table*}[t]
\centering
\begin{tabular}{lrrrrr}
\toprule
\multirow{2}{*}{\textbf{Model}}
& \textbf{TPS} & \textbf{RPS} & \textbf{Avg. Output} & \textbf{Weights on CPU} & \textbf{Peak VRAM} \\
& (token/s) & (req/s) & \textbf{Length} & (GB) & (GB) \\
\midrule
Original & 41.86 & 0.64 & 65.73 & 0 & 1.17 \\
VT (\prealloc) & 41.65 & 0.55 & 64.84 & 0.28 & 0.91 \\
VT (\prealloc) + \diskemb\ & 41.44 & 0.51 & 64.78 & 0 & 0.91 \\
\bottomrule
\end{tabular}
\caption{Throughput and resource utilization for three inference configurations on a machine translation (English-to-Chinese) task ($N=100$), comparing the effect of embedding offloading. Decode throughput is measured in Tokens Per Second (TPS), computed as total generated tokens divided by total decoding time. Request throughput (RPS), average output length, weights on CPU, and peak GPU memory consumption are also reported.}
\label{tab:throughput_stats_lmdb}
\end{table*}

\noindent\textbf{Latency behavior.} Compared to the Original, VT (\prealloc) increases mean TTFT from 1.74 ms to 3.71 ms, reflecting the additional overhead of dynamic LM head construction and CPU-based embedding lookups. Introducing disk-based offloading further increases TTFT slightly to 4.58 ms, as embedding vectors are retrieved from disk rather than CPU memory. Importantly, this overhead remains small in absolute terms.

Tail latency analysis reveals a different trend. While VT (\prealloc) exhibits a higher P99 TTFT (17.20 ms), \diskemb\ configuration reduces P99 TTFT to 12.23 ms, indicating more stable prefill latency. This suggests that disk access introduces predictable overhead without amplifying variance across inputs.

TPOT remains comparable across all configurations, with mean TPOT values clustered around 24 ms. This confirms that embedding offloading---whether to CPU or LMDB---does not affect steady-state decoding, which is dominated by transformer forward passes.

End-to-end (E2E) latency remains broadly consistent across configurations. Despite differences in TTFT, both VT-based models achieve comparable mean and slightly lower tail E2E latency relative to the Original, indicating that prefill overhead is amortized over the generation process for typical output lengths.

\begin{figure*}[ht]
    \centering
    \includegraphics[width=\linewidth]{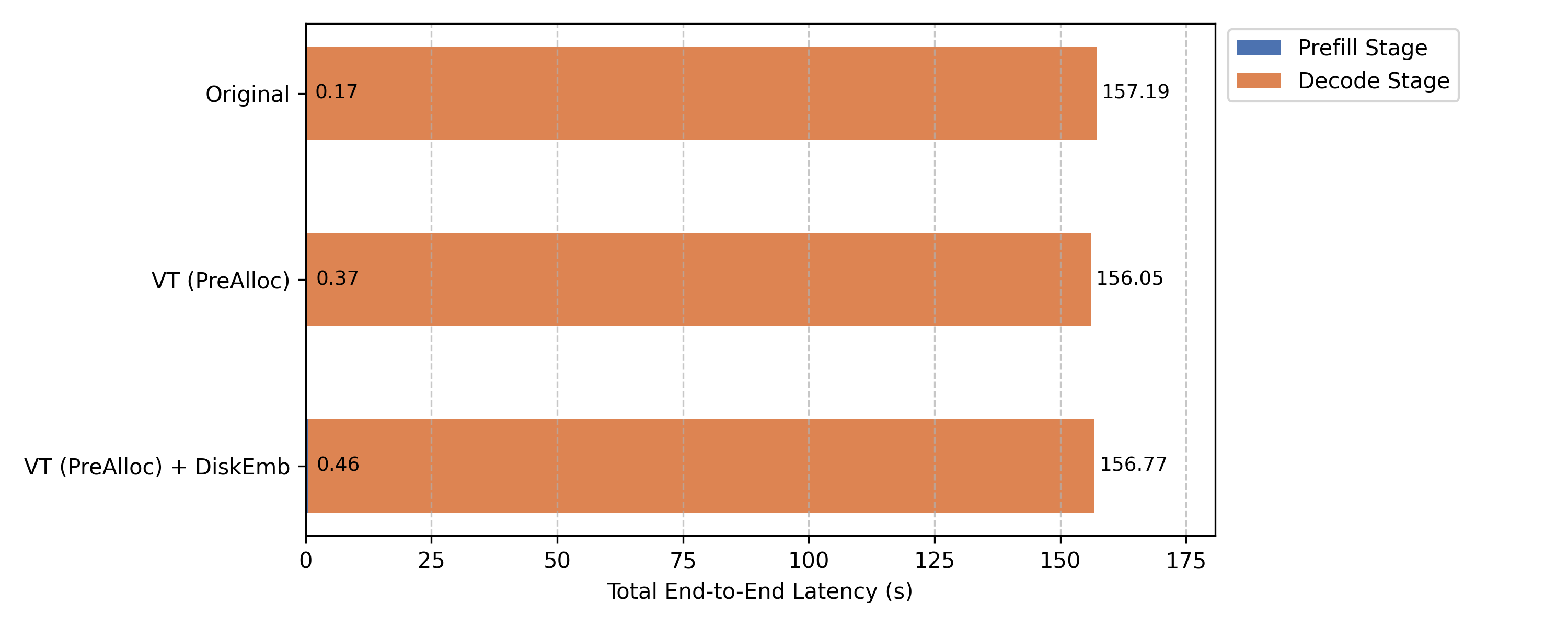}
    \caption{Latency decomposition of three inference configurations. We report the total end-to-end (E2E) latency consumed by the prefill (initialization) stage versus the decoding (generation) stage. While offloading embeddings to the CPU or LMDB increases the prefill proportion relative to the Original ($0.11\%$), the impact remains marginal ($0.24\%$ for VT \prealloc\ and $0.30\%$ for \diskemb), ensuring that the latency remains dominated by decoding.}
    \label{fig:latency_decomp_lmdb}
\end{figure*}

\noindent\textbf{Latency decomposition.} Figure \ref{fig:latency_decomp_lmdb} illustrates the breakdown of total end-to-end (E2E) latency consumed by the prefill stage versus the decode stage. In the vanilla configuration, the prefill stage is nearly instantaneous, accounting for only 0.11\% of total execution time. Transitioning to VT (\prealloc) increases this proportion slightly to 0.24\%. When moving further to \diskemb, the prefill percentage still remains remarkably low at 0.30\%. This demonstrates that while retrieving embedding vectors from disk-based storage (LMDB) is technically more time-consuming than CPU memory access, the impact on the overall latency profile is negligible. In all cases, the decoding stage continues to dominate over 99.7\% of the runtime.

\noindent\textbf{Throughput and resource efficiency}. Table~\ref{tab:throughput_stats_lmdb} reports decode throughput and resource usage. Decode TPS remains stable across configurations (around 42 tokens/s). This supports the observation that offloading embedding does not degrade decoding performance. RPS is slightly lower for VT variants, consistent with their higher per-request initialization overhead. Peak GPU memory is reduced from 1.17 GB (Original) to 0.91 GB for both VT-based configurations, reflecting the benefit of pre-allocating LM head weights on GPU and offloading embeddings. VT (\prealloc) has embedding weights on the CPU, which consumes 0.28 GB of memory. With \diskemb, the embedding is evicted to disk, eliminating the CPU memory overhead.

\noindent\textbf{Summary.} Overall, the results confirm that embedding offloading is compatible with VocabTailor and provides additional memory savings. Both CPU-based and disk-backed offloading improve GPU memory efficiency relative to the original model without compromising decoding performance, making VocabTailor suitable for memory-constrained or resource-limited environments.

\section{Support Both Tied and Non-Tied Embedding Architectures} \label{appx:non-tied}
In large language models, the embedding layer and the LM head share the same weight dimension. The hidden states after going through the LM head and the input tokens before feeding into the transformer layers are considered in a similar representation space. Thus, to reduce the total number of parameters, some models would share the weight of the embedding and the LM head. Weight tying effectively reduces model size and inference-time memory consumption and is therefore widely adopted in small language models (SLMs). However, a portion of SLMs still retains non-tied embeddings to keep higher expressiveness. 



\begin{figure*}[ht]
    \centering
    \includegraphics[width=\linewidth]{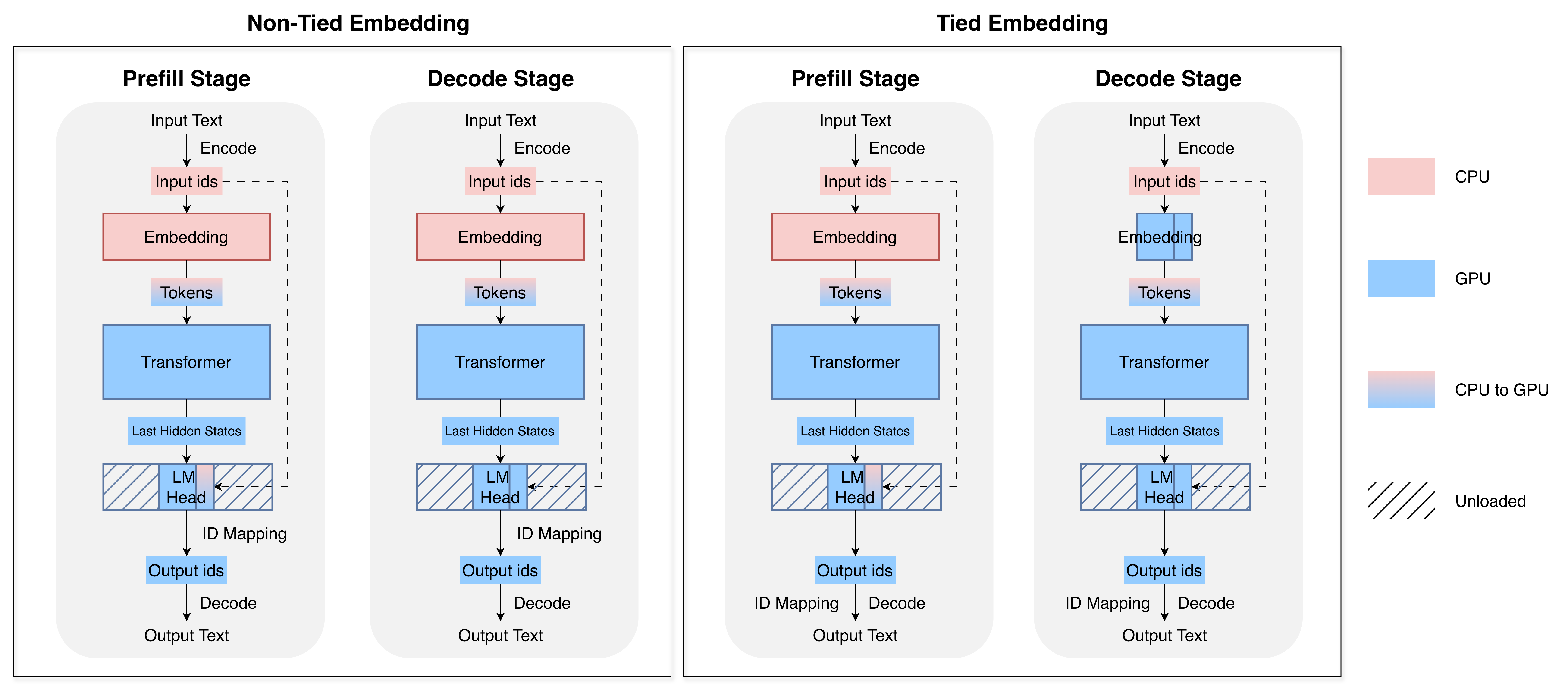}
    \caption{Comparison of VocabTailor workflows. Left: In tied architectures, LM head weights are reused for decoding to eliminate memory movement. ID mapping is deferred until the final tokenizer stage. Right: In non-tied architectures, only the LM head is reduced, so an explicit ID mapping is required after sampling to align the reduced LM head and the full CPU embedding layer.}
    \label{fig:tie_nontied_framework_comp}
\end{figure*}

VocabTailor supports both tied and non-tied embedding architectures, with their respective inference workflows illustrated in Figure~\ref{fig:tie_nontied_framework_comp}. For tied embedding models, token lookup during the prefill stage is performed using the embedding weights stored on the CPU. During decoding, because the embedding and LM head weights are shared, we can directly use the reduced LM head weights to replace the embedding weights without extra memory usage, thereby eliminating memory movement overhead. Moreover, since the LM head is already reduced, embedding lookup can be performed without an ID mapping between the original token index and the reduced version token index. Such mapping is deferred until generation completes, before tokenizer decoding. 

For the non-tied embedding models, the embedding layer remains on the CPU and is used for both prefill and decode stages. Since only the LM head is reduced, the token indices produced by the reduced LM head after the softmax and sampling operation must be mapped back to the original indices before calling the embedding layer forward to get the corresponding token vectors. After generation, the produced token IDs are already the same as the ones in the full vocabulary, so no additional ID mapping is required for the tokenizer decoding.

\section{Profiling Strategy Algorithm} \label{appx:profile_alg}
The detailed profiling strategy is presented in Algorithm~\ref{alg:profiling_strategy}. 

\begin{algorithm}[thp]
\caption{Profiling Strategy}
\label{alg:profiling_strategy}
\textbf{Definition}: \\
$\set{V}$: Corpus-profiling vocabulary \\
$M$: Total number of documents in the corpus \\
$\set{I}_i$: Set of input tokens corresponding to a single example $i$ \\
$\set{U}$: Set of language-specific tokens obtained from Unicode blocks \\
$df \in \mathbb{R}$: Dictionary mapping token $v$ to its document frequency $df(v)$ \\
$\tau \in [0, 1]$: Tolerance threshold (fraction of documents allowed to be impacted) \\
$\set{T}$: Final calibrated task-specific vocabulary \\
\textbf{Input}: $\set{V}$, $\tau$\\
\textbf{Output}: $\set{T}$

\begin{algorithmic}[1] 
\STATE \textbf{Input-Aware Filtering}
\STATE $\set{V}_1 \gets \{ v \in \set{V} \mid v \notin \set{I}_i \}$

\STATE \textbf{Language-specific Filtering}
\STATE $\set{V}_2 \gets \{ v \in \set{V}_1 \mid v \in \set{U} \}$

\STATE \textbf{Tolerance Filtering}
\STATE $N \gets |\set{V}_2|$
\STATE $F \gets \text{list of } (v, df(v)) \text{ for } v \in \set{V}_2$
\STATE Sort $F$ ascending by $df(v)$
\STATE $\code{index} \gets 0$
\FOR{$j \gets 1$ \textbf{to} $N + 1$}
    \IF{\code{sum}($F[:\code{j}]) > \tau M$}
        \STATE $\code{index} \gets N - j + 1$
        \STATE \textbf{break}
    \ENDIF
\ENDFOR
\STATE \RETURN $\set{T} \gets \{ v \in F[-\code{index}:] \}$
\end{algorithmic}
\end{algorithm}

\section{Discussion}
\subsection{Discussion on Memory Reduction}

VocabTailor achieves substantial reductions in the memory usage of vocabulary-related components. In certain tasks (\eg information extraction), this can reduce the active vocabulary to a small fraction of the full set, yielding up to 99\% savings. However, these components (\ie embedding matrix and LM head) account for only a portion of the total model memory, which also includes transformer layers, activations, and KV cache. As a result, the overall reduction in end-to-end VRAM usage is more moderate (22-23\% as shown in Table~\ref{tab:eff_results}, though still meaningful in practice. This distinction highlights the importance of separating component-level savings from system-level memory usage when evaluating efficiency gains.

\subsection{Discussion on Task Switching and Switching Cost}

In practical small language model (SLM) deployment scenarios, models are typically optimized for a single downstream task or a small set of closely related tasks. High-frequency switching between substantially different task-specific vocabularies is relatively uncommon in such settings. In many production settings, SLMs are instantiated as task-specialized services, thereby avoiding repeated reconfiguration during runtime. In scenarios where task switching is required, such switching is generally low-frequency. In these cases, the overhead associated with loading task-specific LM Head parameters remains modest relative to overall deployment and inference costs. For VocabTailor, it takes 0.2 s to switch between task-specific vocabularies, which we consider acceptable in low-frequency scenarios.

\subsection{Discussion on Out-of-distribution}

Out-of-distribution is a common issue for existing vocabulary pruning methods. In fact, one of our contributions is to mitigate this issue. Out-of-distribution is bounded by our design as VocabTailor retains the full tokenizer. (details in Section 3.1), dynamically loads the input tokens for each example at runtime, and maintains a task-specific static vocabulary $T$. Thus, in the prefilling stage, the input text is encoded as the same tokens and generates identical hidden states as the original model without vocabulary pruning. In essence, our design eliminates the out-of-distribution problem for the prefilling stage (which happened for the current static vocab pruning). The only potential limitation concerns rare output-only tokens not captured in $T$. However, tolerance filtering (Section 3.4) explicitly bounds worst-case degradation via $\tau$. In our experiments ($\tau = 0.01$), no catastrophic failure was observed across five diverse tasks (Table 1).

\section{Data Processing and Evaluation Details} \label{appx:data_processing}
For machine translation, we follow the best practice of zero-shot machine translation by \citet{zhang_prompting_2023}, using a simple English template. 

\noindent\textbf{English-to-Chinese translation template:}
\begin{lstlisting}[
    language=Python, 
    breaklines=true,
    breakatwhitespace=true,
    basicstyle=\ttfamily\small
]
[{"role": "user", "content": "English: {SOURCE}\n Chinese:"}]
\end{lstlisting}

\noindent\textbf{English-to-Italian translation template:}
\begin{lstlisting}[
    language=Python, 
    breaklines=true,
    breakatwhitespace=true,
    basicstyle=\ttfamily\small
]
[{"role": "user", "content": "English: {SOURCE}\n Italian:"}]
\end{lstlisting}

For summarization, the model is finetuned and evaluated using prompt shown below. 

\noindent\textbf{Summarization prompt template:}
\begin{lstlisting}[
    language=Python, 
    breaklines=true,
    breakatwhitespace=true,
    basicstyle=\ttfamily\small
]
[{"role": "user", "content": f"Document:\n{DOCUMENT}\n"}]
\end{lstlisting}



For the code completion task, we use the evaluation script of SAFIM. For information extraction, we use the LM-Eval-Harness framework and set the task to `squad\_completion`. For math word problem solving, we use the Math-Eval-Harness framework and set the task to `mawps`.

\section{Fine-tuning Config for Summarization Baseline Model} \label{appx:fine_tune}
For summarization, all the methods are based on a finetuned Llama 3.2 3B model. We set learning rate=2e-5, batch size = 128. The model is fine-tuned using XSUM train split set for 1 epoch. 

\section{Use of AI for Writing/Coding Assistance} \label{appx:ack}
Generative AI use in this work is limited to assistance purely with the language of the paper and short-form input assistance under the AI writing assistance policy.

\end{document}